\documentclass[11pt]{article}
\usepackage{amssymb,amsmath,bm,graphicx}
\usepackage{mathrsfs}
\usepackage{constants}
\usepackage{enumitem}
\usepackage{textcomp}
\usepackage{url}
\usepackage{verbatim}
\usepackage{cite}
\usepackage{xcolor}
\usepackage{microtype}
\usepackage{longtable}
\usepackage{tabto}
\usepackage{amsmath}
\usepackage{float}
\usepackage{subcaption}
\usepackage{caption}
\usepackage{algpseudocode}
\usepackage{algorithm}
\usepackage{color}

\usepackage{float}

\usepackage{placeins}

\usepackage{amssymb}
\usepackage{graphicx}

\usepackage{subcaption}
\usepackage{caption}

\usepackage{csvsimple}
\usepackage[T2A]{fontenc}
\usepackage[utf8]{inputenc}
\usepackage{babel}
\usepackage{dblfloatfix}
\usepackage{amsmath}
\usepackage{lipsum}
\usepackage{amsfonts}
\usepackage{caption}

\usepackage{newtxmath,booktabs,sectsty}
\paragraphfont{\mdseries\itshape}
\usepackage{tabularx,ragged2e}
\newcolumntype{L}[1]{>{\RaggedRight\hsize=#1\hsize}X}
\newcolumntype{C}[1]{>{\Centering\hsize=#1\hsize\hspace{0pt}}X}

\date{}
\begin{document}
\newcommand{\bea}{\begin{eqnarray}}
\newcommand{\ena}{\end{eqnarray}}
\newcommand{\beas}{\begin{eqnarray*}}
\newcommand{\enas}{\end{eqnarray*}}
\newcommand{\beq}{\begin{equation}}
\newcommand{\enq}{\end{equation}}
\def\qed{\hfill \mbox{\rule{0.5em}{0.5em}}}
\newcommand{\bbox}{\hfill $\Box$}
\newcommand{\ignore}[1]{}
\newcommand{\ignorex}[1]{#1}
\newcommand{\wtilde}[1]{\widetilde{#1}}
\newcommand{\qmq}[1]{\quad\mbox{#1}\quad}
\newcommand{\qm}[1]{\quad\mbox{#1}}
\newcommand{\nn}{\nonumber}
\newcommand{\Bvert}{\left\vert\vphantom{\frac{1}{1}}\right.}
\newcommand{\To}{\rightarrow}
\newcommand{\E}{\mathbb{E}}
\newcommand{\Var}{\mathrm{Var}}
\newcommand{\Cov}{\mathrm{Cov}}
\newcommand{\Corr}{\mathrm{Corr}}
\newcommand{\dist}{\mathrm{dist}}
\newcommand{\diam}{\mathrm{diam}}
\makeatletter
\newsavebox\myboxA
\newsavebox\myboxB
\newlength\mylenA
\newcommand*\xoverline[2][0.70]{%
    \sbox{\myboxA}{$\m@th#2$}%
    \setbox\myboxB\null
    \ht\myboxB=\ht\myboxA%
    \dp\myboxB=\dp\myboxA%
    \wd\myboxB=#1\wd\myboxA
    \sbox\myboxB{$\m@th\overline{\copy\myboxB}$}
    \setlength\mylenA{\the\wd\myboxA}
    \addtolength\mylenA{-\the\wd\myboxB}%
    \ifdim\wd\myboxB<\wd\myboxA%
       \rlap{\hskip 0.5\mylenA\usebox\myboxB}{\usebox\myboxA}%
    \else
        \hskip -0.5\mylenA\rlap{\usebox\myboxA}{\hskip 0.5\mylenA\usebox\myboxB}%
    \fi}
\makeatother

\newtheorem{theorem}{Theorem}[section]
\newtheorem{corollary}[theorem]{Corollary}
\newtheorem{conjecture}[theorem]{Conjecture}
\newtheorem{proposition}[theorem]{Proposition}
\newtheorem{lemma}[theorem]{Lemma}
\newtheorem{definition}[theorem]{Definition}
\newtheorem{example}[theorem]{Example}
\newtheorem{remark}[theorem]{Remark}
\newtheorem{case}{Case}[section]
\newtheorem{condition}{Condition}[section]
\newcommand{\proof}{\noindent {\it Proof:} }

\title{{\bf\Large 4TaStiC: Time and trend traveling time series clustering for classifying long-term type 2 diabetes patients}}

\author{Onthada Preedasawakul\thanks{This author is supported by Petchra Pra Jom Klao Master’s Degree Research Scholarship from King Mongkut’s University of Technology Thonburi, Grant 
number: 32/2567. Email: o.preedasawakul@gmail.com} \ and Nathakhun Wiroonsri\thanks{This author is supported by KMUTT Partnering Initiative Grant, fiscal year 2024 under KIRIM number 28105. Email: nathakhun.wir@kmutt.ac.th}  \\ Statistics, Probability, and Data Science with R programming (SPiD$\epsilon$R) research group \\ Department of Mathematics, King Mongkut's University of Technology Thonburi}


\maketitle

\begin{abstract}

Diabetes is one of the most prevalent diseases worldwide, characterized by persistently high blood sugar levels, capable of damaging various internal organs and systems. Diabetes patients require routine check-ups, resulting in a time series of laboratory records, such as hemoglobin A1c, which reflects each patient's health behavior over time and informs their doctor's recommendations. Clustering patients into groups based on their entire time series data assists doctors in making recommendations and choosing treatments without the need to review all records. However, time series clustering of this type of dataset introduces some challenges; patients visit their doctors at different time points, making it difficult to capture and match trends, peaks, and patterns. Additionally, two aspects must be considered: differences in the levels of laboratory results and differences in trends and patterns. To address these challenges, we introduce a new clustering algorithm called Time and Trend Traveling Time Series
Clustering (4TaStiC), using a base dissimilarity measure combined with  Euclidean and Pearson correlation metrics. We evaluated this algorithm on artificial datasets, comparing its performance with that of seven existing methods. The results show that 4TaStiC outperformed the other methods on the targeted datasets. Finally, we applied 4TaStiC to cluster a cohort of 1,989 type 2 diabetes patients at Siriraj Hospital. Each group of patients exhibits clear characteristics that will benefit doctors in making efficient clinical decisions. Furthermore, the proposed algorithm can be applied to contexts outside the medical field.

\end{abstract}




\section{Introduction}\label{sec1}

Diabetes is a chronic metabolic condition characterized by persistently high blood sugar levels, which over time are capable of damaging vital organs such as the heart, blood vessels, eyes, kidneys, and nerves. Among the various types of diabetes, type 2 remains the most common worldwide \cite{khan2020}. The World Health Organization (2024) \cite{WHO2024} reports that over 800 million adults are currently affected by diabetes globally, with its prevalence almost doubling from approximately 7\% in 1990 to nearly 14\% in 2022. This dramatic increase highlights diabetes as a significant and growing public health concern. Diabetic retinopathy is a common and serious complication of type 2 diabetes, representing a leading cause of vision loss and the second most common cause of blindness after cataracts \cite{DR2}. Because the early stages of diabetic retinopathy are often asymptomatic, routine screening and patient stratification are essential for early intervention \cite{DR2_prevent}.
Type 2 diabetes has a highly heterogeneous presentation \cite{diabetes_het}, exhibiting a broad range of clinical trajectories, comorbidities, and treatment responses. This variability challenges one-size-fits-all treatment strategies, making the stratification of patients into subgroups with similar clinical characteristics a crucial step towards precision medicine \cite{precisionmed1}.

For routine clinical visits, doctors typically focus on reviewing a limited selection of each patient's recent records to inform their recommendations. However, the entire time series of records is meaningful and can reflect patients' behaviors over time. Intensive examination of each patient's whole record is not practical, especially in the context of the Thai public medical system \cite{ThaiHS}, which serves a great number of patients \cite{ThaiHS2} due to easy access to specialists. The unsupervised classification of patients with similar characteristics, extracted from the entirety of their time series records, would help to guide doctors towards more precise and effective recommendations. To this end, we collaborated with Siriraj Hospital, one of the largest and most prestigious hospitals in Thailand \cite{SI}, to attempt to classify their diabetes patients using their hemoglobin A1c (HbA1c) time series data. If successful, this research would lead to the creation of a dashboard plugged-in into the Siriraj system that would display each patient's group and guideline recommendation. 

According to the Centers for Disease Control and Prevention (CDC), an HbA1c between 5.7\% to 6.4\% and an HbA1c of $\ge$6.5\%  are diagnosed as prediabetes and diabetes, respectively. Patients' visit times and gaps between visits vary, yet they should not have an impact on the clustering outcomes, presenting a challenge. For instance, if there are two patients with declining HbA1c trends, one consistent and the other highly fluctuating, they should be assigned to different groups. However, two patients with slightly different HbA1c levels and slightly different declining trends but similar fluctuating patterns may be classified into the same group. This aim of this work is to develop a method to solve these challenges and cluster patients based on both HbA1c levels and hidden trends and patterns.

Cluster analysis \cite{clustering} is a widely used unsupervised machine learning technique for partitioning data into related groups based on their characteristics, with applications in various fields (see \cite{bio1,bio2,xue2017joint,xu2017novel}).  In the context of temporally ordered data, this technique is referred to as time-series clustering. Time series clustering is an unsupervised approach for grouping a time series dataset into distinct clusters where each cluster comprises series that display similar patterns or behaviors across multiple time points, even in the presence of noise, amplitude differences, or local temporal misalignments \cite{timeclust}. It has been used in diverse fields such as psychology, finance, and electrical engineering (see, for example \cite{elecmonitoring, powerconsumption, timefinance}). In the medical field, each patient's records—such as weight and laboratory results—are collected over time, making time series clustering essential for capturing temporal dynamics and evolving patterns \cite{timepattern}. Time series clustering has been used in several healthcare studies (see, for instance, \cite{ex_timeclust1,ex_timeclust2,ex_timeclust3,ex_timeclust4,mpox}, and references therein). However, clustering time series data using medical records, such as diabetic patients' HbA1c levels, presents unique challenges. While records at different time points may seem unrelated when viewed statically, they may exhibit similar trends and patterns with temporal lags. Partial solutions are offered by traditional methods, such as dynamic time warping (DTW) \cite{dtw1, dtw2}, global alignment kernel (GAK) \cite{gak}, and cross-correlation similarity measures \cite{autocorr}. DTW excels with handling time shifts but can over-align unrelated sequences. GAK enables global alignment but lacks local variability. Autocorrelation captures periodicity but ignores timing mismatches. A more recent technique, lag penalized weighted correlation (LPWC) \cite{lpwc}, penalizes over-flexible alignments but is too strict for this problem and does not account for trend heterogeneity.

Building upon the above discussions, we introduce a novel dissimilarity measure named time and trend traveling time series clustering (4TaStiC). 4TaStiC computes the dissimilarity by shifting time points and slightly tilting trends to find the best time and trend match between two time series. A combination of Euclidean and Pearson correlation metrics is used for the base dissimilarity. Then, a clustering algorithm, such as hierarchical clustering \cite{hc1,hc2}, DBSCAN \cite{DBSCAN}, or OPTICS \cite{OPTICS}, receives and processes the dissimilarity matrix to find the final clusters. 

\begin{figure*}[hbt!]
\centering
\resizebox{1.0\textwidth}{!}{%
\begin{tabular}{cc}
\centering\includegraphics[width=4cm]{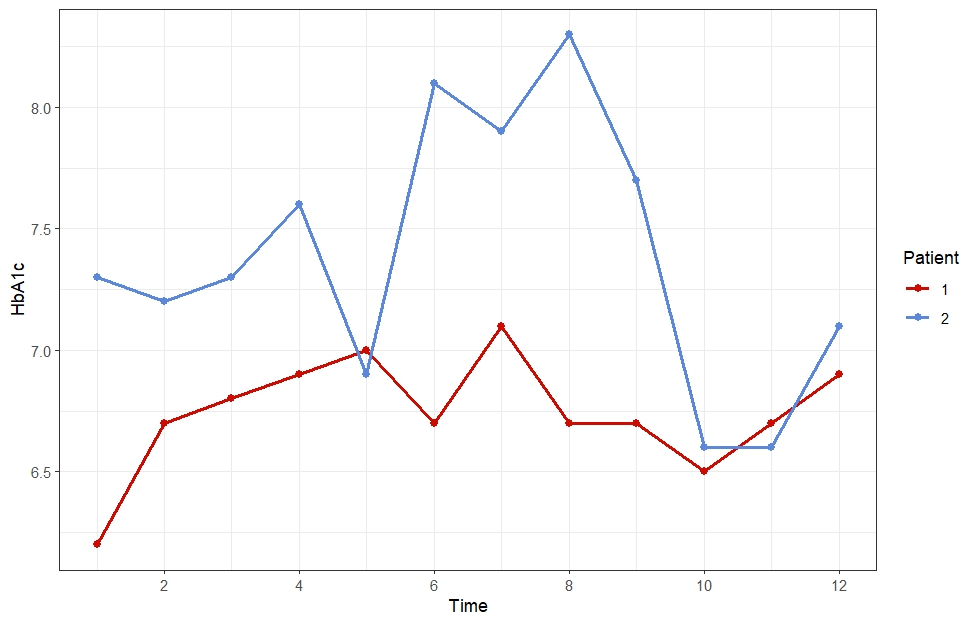} & \includegraphics[width=4cm]{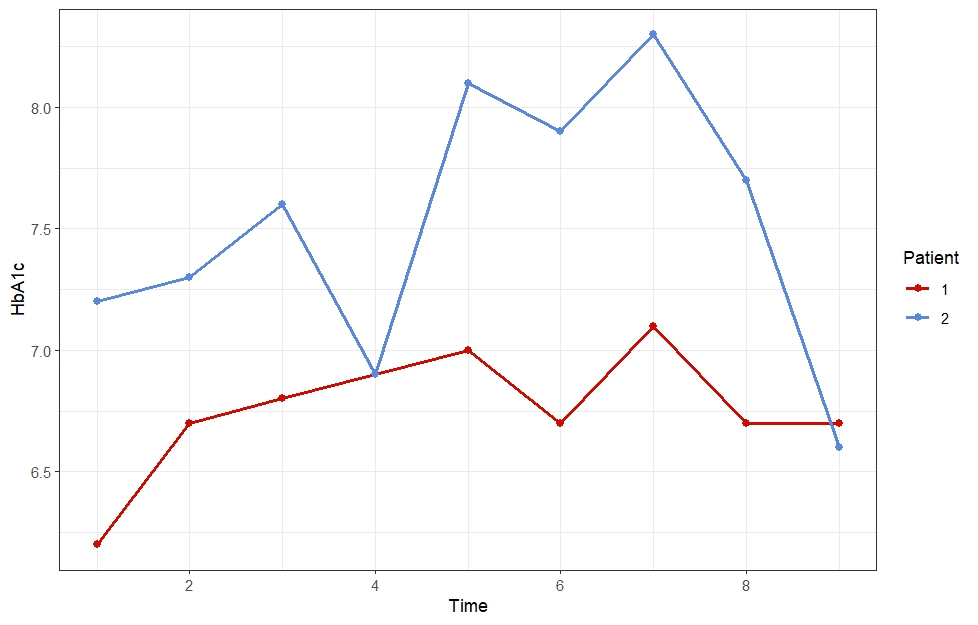} \\
\centering\includegraphics[width=4cm]{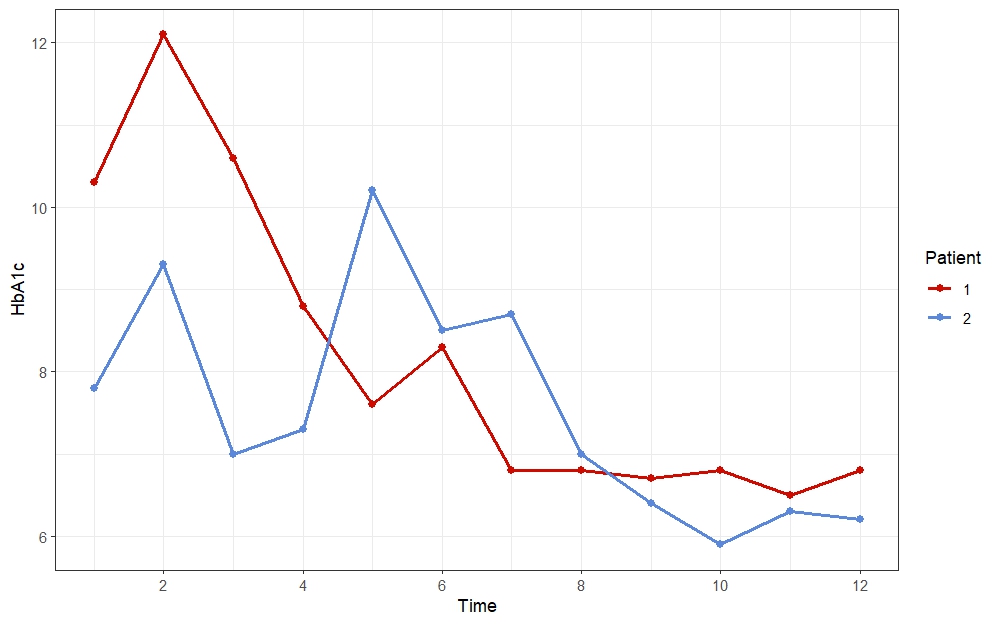} & \includegraphics[width=4cm]{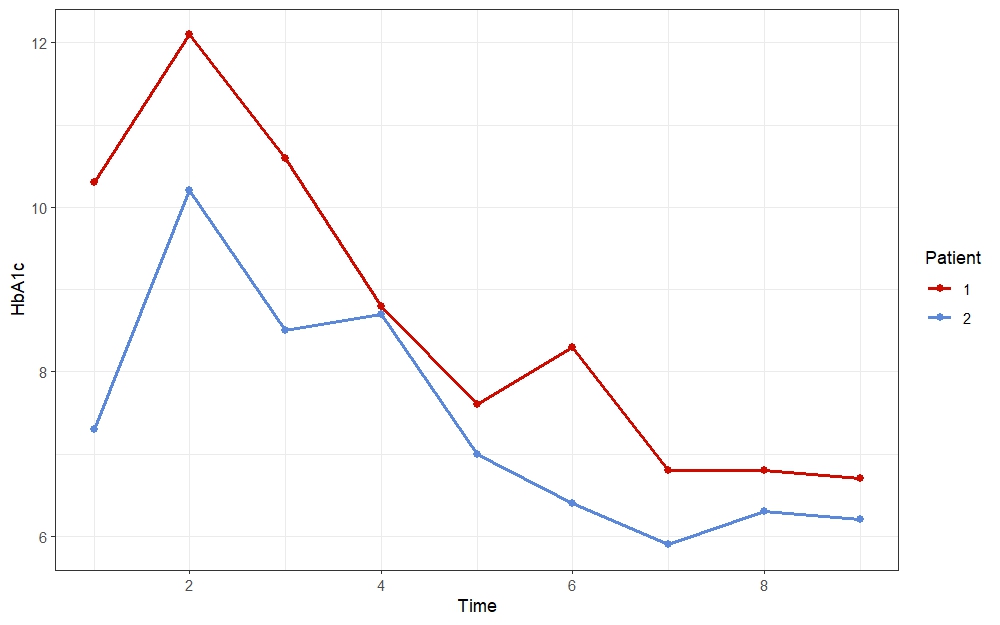} \\
\end{tabular}
}
\caption{Examples of two pairs of patients' HbA1c time series before (left) and after time and trend traveling (right) }
\label{fig:ex1}
\end{figure*}

Recognizing the inherent complexities of medical datasets, particularly diabetes patient records, we integrate the advantages of both time traveling and trend traveling approaches. Medical data frequently exhibit discrepancies due to irregular appointment schedules and slight variations in patients' temporal trends (see Figure \ref{fig:ex1} for instance). Traditional dissimilarity may struggle to group or separate patients whose patterns differ in terms of timing or trends. By combining time-shifting and subtle trend-aligning rotations, 4TaStiC constructs a comprehensive dissimilarity measure specifically tailored to the nuanced temporal structure of medical data. This integrated method ensures the accurate grouping of patients who share intrinsic patterns or behavior while carefully differentiating those whose patterns genuinely diverge. Although we are motivated by medical data, 4TaStiC can be applied to all domains in which it is necessary to identify similar patterns with varying time points.

The remainder of this work is organized as follows. Section \ref{sec:overview} opens with the model overview and some necessary background. Our proposed methodology is stated and discussed in Section \ref{sec:methods}. Section \ref{sec:exp} shows the experimental results on artificial data. Section \ref{sec:app} is devoted to an application to diabetes patients' data. Finally, a discussion and future research directions are presented in Section \ref{sec:discuss}.

\section{Overview and background} \label{sec:overview}

This section gives our model overview and states and discusses all the terms and existing methods used in this work.

\subsection{Our proposed model overview}

Time series clustering is a well-known technique that includes many large existing algorithms; however, most of these algorithms focus on differentiating observations by either their physical distances or their behaviors, but not both. In addition, they struggle to handle cases in which time points and time gaps are different. Furthermore, the practice of slightly shifting time series trends to emphasize clearer behaviors has never been considered. The method we propose is intended to resolve all of these issues. Specifically, we integrate the Euclidean distance and the correlation-based dissimilarity. The idea of combining two metrics has been studied before (see, for instance, \cite{RDPC2025}). Then, when finding the distance between two time series, we search for the best match by shifting the time for a predetermined maximum number of time steps to be shifted and tilting the trend for epsilon-small angles. Our proposed dissimilarity algorithm can be naturally attached to existing clustering algorithms such as hierarchical clustering, DBSCAN, OPTICS, etc. In this work, we focus on hierarchical clustering. This is the time and trend traveling time series clustering approach (4TaStiC) detailed later in Section \ref{sec:methods}. It is appropriate for handling cases in which there is uncertainty about starting times and time gaps and there is a need to maximize the opportunity to detect similar patterns, even if they exhibit slightly different trends. 

\begin{figure*}[t]
\centering\includegraphics[width=12cm]{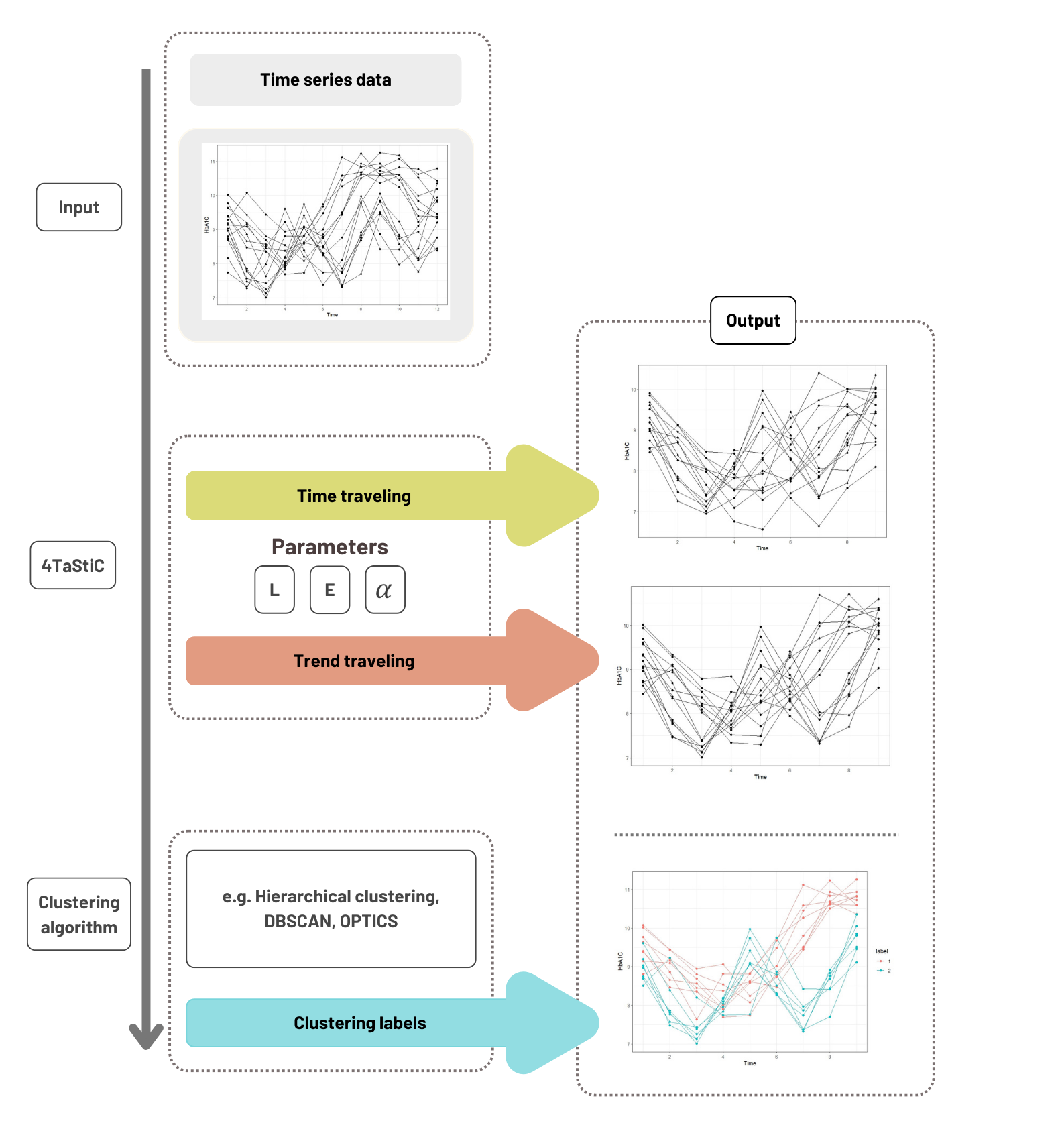}
\caption{The 4TaStiC idea map for diabetes patients}
\label{fig:flow4}
\end{figure*}

We control the weight between the Euclidean distance and the correlation-based dissimilarity, and the time and trend traveling levels for 4TaStiC using the following four predetermined parameters: 
\begin{itemize}
    \item $\alpha$ is a numerical value between 0 and 1 that represents the proportions of the Euclidean distance and the correlation-based dissimilarity to be computed. A larger $\alpha$ increases the importance of correlation-based dissimilarity.
    \item $L$ is a non-negative integer denoting the maximum time steps to be shifted during the 4TaStiC calculation.
    \item $E$ is a set of real numbers representing angles to be tilted when computing 4TaStiC. $E$ should contain 0 and a few other numbers close to zero.
    \item $C \ge 0$ is a penalty coefficient that reduces the impact of the correlation term when tilting time series data. Its function is to avoid a problem with excessively large tilting angles. 
\end{itemize}

The 4TaStiC idea workflow is illustrated in Figure \ref{fig:flow4}. All calculations were performed within the RStudio environment \cite{RStudio}. The “dtwclust” \cite{dtwclust} and “LPWC” \cite{Rlpwc} packages were utilized to facilitate comparisons between our proposed method and existing approaches.

\subsection{Clustering methods}
\subsubsection{K-means Clustering \cite{Kmeans1,kmeans2}}
K-means is a partition-based clustering method that divides a dataset into $K$ clusters and aims to minimize the squared error between each data point and its corresponding cluster centroid, defined as 
\beas \label{kobj}
J(K) = \sum_{j=1}^{K} \sum_{x \in C_j} d^2_{eu}(x,v_j),
\enas
where $C_j$ denotes the set of  data points in the $j^{th}$ cluster and $d_{eu}$ is the Euclidean distance defined later.

\subsubsection{Hierarchical clustering \cite{hc1,hc2}}
Hierarchical clustering is used to group similar data points into clusters based on their similarity. It constructs a nested hierarchy of clusters using either an agglomeration or a divisive approach. In this work, we will focus on the agglomeration approach.

For agglomerative hierarchical clustering, each data point initially forms its own cluster. The algorithm iteratively merges the two closest clusters based on a specified linkage criterion until either all data points form a single cluster or a predetermined condition is reached. The well-known linkages are single, complete, and average, as defined in Table \ref{linkage}.

\begin{table}[h]
    \centering
    \begin{tabular}{lll}
        \hline\hline
        \textbf{Linkage} & \textbf{Description} & \textbf{Formula} \\
        \hline
        \textbf{Single} 
        & The closest pair of points
        & $\min\limits_{x \in C_i, y \in C_j} d(x,y)$ \\
        \textbf{Complete} 
        & The farthest pair of points
        & $\max\limits_{x \in C_i, y \in C_j} d(x,y)$ \\
        \textbf{Average} 
        &The average distance between all points 
        & $\frac{1}{|C_i||C_j|} \sum\limits_{x \in C_i}\sum\limits_{y \in C_j}d(x,y)$ \\
        \hline\hline
    \end{tabular}
\caption{\textbf{Agglomerative clustering linkages}}
\label{linkage}
\end{table}

\subsection{Cluster evaluation} \label{evaluation}
To assess the quality of clustering results, we use the ordinary accuracy and the ARI \cite{ARI}, which is a widely used external validation metric. The ARI evaluates the agreement between the predicted clustering and ground truth labels while adjusting for chance grouping. The ARI is defined as:

\begin{equation*}
ARI (A,C) = \frac{
    \sum_{ij} \binom{|A_i\cap C_j|}{2} - \left[ \sum_i \binom{|A_i|}{2} \sum_j \binom{|C_j|}{2} \middle/ \binom{n}{2} \right]
    }
{
\frac{1}{2} \left[ \sum_i \binom{|A_i|}{2} + \sum_j \binom{|C_j|}{2} \right] - \left[ \sum_i \binom{|A_i|}{2} \sum_j \binom{|C_j|}{2} \middle/ \binom{n}{2} \right]
}
\end{equation*}
where $A = \{A_1, A_2, \dots, A_k\}$ is the set of $k$ clusters obtained from the algorithm, $C = \{C_1, C_2, \dots, C_K\}$ is the true partition with $k,K \in \mathbb{N}$, and $n$ is the total number of data points. The ARI ranges from $-1$ to $1$, where the value of $1$ indicates the perfect agreement.

\subsection{Existing distances and dissimilarity measures}
Dissimilarity measures play a crucial role in clustering by characterizing the differences between data points. The choice of distance metric or dissimilarity measure significantly affects the clustering outcomes, particularly for time series data where capturing temporal information is indispensable. This section provides an overview of commonly used dissimilarity metrics for time series clustering, as summarized in Table \ref{tab:dissimilarity}. This is not the full list, but includes those used for the comparison test in Subsection \ref{performance}. Note that for $T \in \mathbb{N}$, we write $X = (x_1,x_2,\ldots,x_T)$, $Y = (y_1,y_2,\ldots,y_T) \in \mathbb{R}^T$ and the notations $X_{(l^+,0)}$ and $X_{(l^-,0)}$ are defined later in the Section \ref{sec:methods}.

\begin{table}[hp]
\centering
\includegraphics[width=14cm]{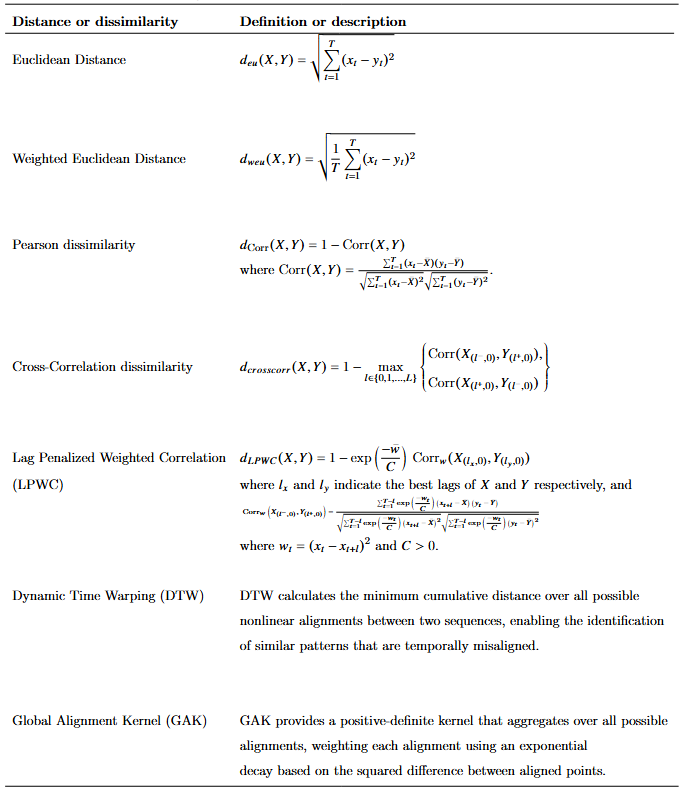}
\caption{Summary of distances and dissimilarity measures for time series clustering used in this work. Since DTW and GAK are algorithmic, they are described conceptually.}
\label{tab:dissimilarity}
\end{table}

\section{Our proposed methods} \label{sec:methods}

In this section, we state our proposed time series clustering algorithm, along with its mathematical properties.

We first define some notations used in this section. For $T \in \mathbb{N}$, we write $X = (x_1,x_2,\ldots,x_T) \in \mathbb{R}^T$ to represent a time series. Letting $l \in \mathbb{N}_0$ such that $l<T$ and $\epsilon \in \mathbb{R}$, we denote

\bea \label{rightshift}
X_{(l^+,\epsilon)} = \left(x_1, x_2 + \epsilon, x_3 + 2\epsilon, \ldots, x_{T-l} + ((T-l)-1)\epsilon \right)
\ena
and 
\bea \label{leftshift}
X_{(l^-,\epsilon)} = \left(x_{l+1}, x_{l+2} + \epsilon, x_{l+3} + 2\epsilon,\ldots,x_T +  ((T-l)-1)\epsilon\right).
\ena

Below, we discuss these notations in depth in specific cases where $\epsilon = 0$, $l=0$, and $l>0$ $\epsilon \in \mathbb{R}$, which are called time traveling, trend traveling, and time and trend traveling, respectively. From this point, $d(\cdot,\cdot)$ is an arbitrary distance or dissimilarity measure.

\subsection{Time traveling}
Lag time in time series refers to a temporal delay between two events. This concept is important for handling medical laboratory results, as pointed out in the introduction. Several existing methodologies are introduced to handle this challenge, such as DTW, GAK, cross-correlation, and LPWC. DTW and GAK use the warping concept, which does not fit with our medical application. Cross-correlation is defined similarly to the time traveling concept, where both solve for the best match for each pair of time series data points, except that it is only defined for correlation-based dissimilarities. LPWC, on the other hand, searches for the best lags for the entire dataset with penalties applied when lags are used, and computes the dissimilarity matrix of the new shifted dataset. In this work, we search for the best match for each pair of data points regardless of what distance or dissimilarity is used.

This concept can be summarized in terms of \eqref{rightshift} and \eqref{leftshift}, with $\epsilon=0$ or equivalently $E=\{0 \}$. Let $L<T$ be a positive integer representing the maximum number of time shifts to be considered. The time traveling measure is defined as 

\bea \label{TimeT}
d_{L}(X,Y) = \min_{l \in \{0,1,\ldots,L\}}\{d(X_{(l^-,0)},Y_{(l^+,0)}),d(X_{(l^+,0)},Y_{(l^-,0)})\}.
\ena

By taking the minimum, we compute the distance at the best time match. It is clear that the notations $l^+$ and $l^-$ refer to forward and backward time shifts, respectively. Again, the cross-correlation dissimilarity is defined similarly to \eqref{TimeT} with $d(\cdot,\cdot)$ representing the Pearson correlation dissimilarity. No penalty terms are introduced here since we aim to find the best time match such that shifting size does not affect the outcome. In this work, we apply it to a combination of Euclidean distance and the Pearson correlation dissimilarity, combined with the new trend traveling technique discussed below.

\subsection{Trend traveling}

In this subsection, we propose a novel concept called trend traveling, which arises from our observation that some pairs of patients exhibit almost identical behavioral patterns, while exhibiting slightly different overall trends over time. In such cases, traditional correlation measures may be smaller than we expect. Tilting the time series data with a tiny angle before computing the correlation can help capture those similar patterns. In general, let $E \subset \mathbb{R}$ be a finite set of the tilting parameters, then the trend traveling is defined as

\bea \label{TrendT}
d_E(X,Y) = \min_{\epsilon \in E}\{d(X_{(0,0)},Y_{(0,\epsilon)})\}.
\ena

Note that the shifting trend is unreasonable when $d(\cdot,\cdot)$ is any physical distance. It should only be applied to correlation-based dissimilarities. 

Nevertheless, an excessive rotation can introduce artificial distortions that violate the real meaning of the data. A penalty term $e^{-C| \epsilon |}$ is introduced to solve the problem
where $C\ge 0$ determines the level of penalty and $\epsilon$ is a tilting angle as in \eqref{rightshift} and \eqref{leftshift}. It is clear that a larger tilting angle results in a greater penalty. For instance, it can be added to the Pearson correlation dissimilarity as

\bea \label{trendpenalty}
d(X_{(0,0)},Y_{(0,\epsilon)}) =1- e^{-C| \epsilon |}  \Corr(X_{(0,0)},Y_{(0,\epsilon)}),
\ena

where $\Corr(\cdot,\cdot)$ is the Pearson correlation, as shown in Table \ref{tab:dissimilarity}. We assume conventionally throughout the paper that $\Corr(0,0) = 1$.

\subsection{4TaStiC dissimilarity measure}
Building upon our earlier discussion, we introduce the 4TaStiC dissimilarity measure. The 4TaStiC measure integrates two concepts, as discussed in the previous subsections: time traveling and trend traveling. By quantifying similarity through temporal time shifts and trend alignments, the 4TaStiC method rigorously identifies time series patterns that exhibit similar temporal behaviors. This proposed method ensures that time series exhibiting closely aligned temporal behaviors are grouped effectively, while those with differing temporal patterns are distinctly separated.

Letting $L$ and $E$ be the same as in the previous two subsections, we define the 4TaStiC metric as 

\begin{equation} \label{4TaStiC}
\begin{aligned}
d_{L,E}(X,Y) = \min_{\substack{l \in \{0,1,\ldots,L\} \\ \epsilon \in E}} \left\{ d(X_{(l^-,0)},Y_{(l^+,\epsilon)}),  d(X_{(l^+,0)},Y_{(l^-,\epsilon)}) 
\right\}.
\end{aligned}
\end{equation}

In this work, we specifically consider $d(\cdot,\cdot)$ to be the interpolation of the correlation-based distance with the penalty term as defined in \eqref{trendpenalty} and the weighted Euclidean distance defined in Table \ref{tab:dissimilarity}. Specifically, for some pre-selected $C\ge 0$, we let

\bea \label{basedist}
d\left(X_{(l^-,0)},Y_{(l^+,\epsilon)}\right) = \alpha \left(1 - e^{-C| \epsilon |}\Corr(X_{(l^-,0)},Y_{(l^+,\epsilon)})\right) \nn \\ + (1-\alpha)d_{weu}\left(X_{(l^-,0)},Y_{(l^+,0)}\right).
\ena

The second component in \eqref{4TaStiC} is defined similarly, by switching the roles of $l^+$ and $l^-$. The trend traveling parameter is not applied to the second term of \eqref{basedist}, regardless of the value of $\epsilon$, since tilting is not reasonable when computing any physical distance. 

Later, in the diabetes patients application in Section \ref{sec:app}, we set $L=3$, $E=\{-0.075,0,0.075 \}$, and $C=0$. These parameters are selected based on our sensitivity analysis in Subsection \ref{sen_test}.

\subsection{4TaStiC clustering algorithm}
To incorporate the 4TaStiC dissimilarity measure into a clustering algorithm, we first compute a dissimilarity matrix using \eqref{4TaStiC} based on pre-selected parameters, and subsequently perform a clustering algorithm such as hierarchical clustering, DBSCAN, or OPTICS. In this work, we focus on hierarchical clustering.
The algorithm is summarized in Algorithm \ref{alg:alg1}. To make it available to everyone, we built our package called ``FourTaStiC,'' available on \url{https://github.com/nwiroonsri/FourTaStiC} within the RStudio environment  \cite{RStudio}.

\begin{algorithm}[h!]
\caption{4TaStiC Clustering}
\label{alg:alg1}
\begin{algorithmic}
\State \textbf{Input:} A time series dataset $\left(X^{(1)},\ldots,X^{(n)}\right)$, $L$, $E$, $C$, $\alpha$ (default as in \eqref{alpha}), K, clustering method (hierarchical clustering or DBSCAN or OPTICS)
\State \textbf{Output:} 4TaStiC dissimilarity matrix (D), clustering labels
\State \textbf{Define} an initial dissimilarity matrix $D = \left[d\left(X^{(i)}, X^{(j)}\right) \right]_{1\le i \le j \le n}$.

\For{each pair $X^{(i)}$ and $X^{(j)}$ where $1\le i<j\le n$}
\For{$l \in \{1, \ldots, L\}$}
 \If{$0 \in E$}
 \State  right shift: $d_1$ $\gets d\left(X^{(i)}_{(l^-,0)}, X^{(j)}_{(l^+,0)}\right)$
 \State  left shift: $d_2$ $\gets d\left(X^{(i)}_{(l^+,0)}, X^{(j)}_{(l^-,0)}\right)$
 \State $D_{ij} = \min\{D_{ij},d_1,d_2\}$
 \EndIf
 \For{$\epsilon \in E\backslash \{ 0\}$}
 \State  tilt without shift: $d_3$ $\gets d\left(X^{(i)}_{(0,0)}, X^{(j)}_{(0,\epsilon)}\right)$
 \State  tilt with right shift: $d_4$ $\gets d\left(X^{(i)}_{(l^-,0)}, X^{(j)}_{(l^+,\epsilon)}\right)$
 \State  tilt with left shift: $d_5$ $\gets d\left(X^{(i)}_{(l^+,0)}, X^{(j)}_{(l^-,\epsilon)}\right)$
 \State $D_{ij} \gets \min\{D_{ij}, d_3, d_4, d_5\}$
 \State $D_{ji} \gets D_{ij}$
      \EndFor
    \EndFor
  \EndFor
\State Apply a selected clustering algorithm using $D$
\State \textbf{return} The final dissimilarity matrix $D$, the final clustering labels vector
\end{algorithmic}
\end{algorithm}

Next, we discuss how to select all the parameters. The parameters $L$, $E$, and $C$ are based on users' beliefs and experiences with specific applications; however, we recommend $L \le 3$ and $E$ as a set of three numbers $\{-\epsilon,0,\epsilon\}$ for some $\epsilon >0$ due to the time complexity. $C \ge 0$ can be chosen to avoid the scenario where we accidentally select an excessively large $\epsilon$. $K$ may be guided by the elbow method as discussed earlier. The parameter $\alpha \in [0,1]$ is weighted, balancing between the Pearson correlation dissimilarity and the Euclidean distance. The parameter $\alpha$ is flexible and can be selected by users; however, in this work, we use

\begin{equation} \label{alpha}
\alpha = \frac{
\displaystyle \max_{i \ne j} \{d_{eu}(X^{(i)}, X^{(j)})\}
}{
Q_p \left( \{1 - \Corr(X^{(i)}, X^{(j)}) \}_{i \ne j}\right) + \displaystyle \max_{i \ne j} d_{eu}(X^{(i)}, X^{(j)})
}
\end{equation}
where $X^{(1)},X^{(2)},\ldots,X^{(n)}$ with $n \in \mathbb{N}$ are all the time series we aim to cluster, and $Q_p$ denotes the $p^{th}$ percentile. This is chosen to ensure that the Euclidean term does not dominate the 4TaStiC when the scale of $X$ we consider is large. It is obvious that a smaller $p$ value results in a larger weight on the Pearson correlation dissimilarity.

\subsection{Mathematical properties}

We complete this section by stating and proving some properties of 4TaStiC. The following proposition shows that the 4TaStiC dissimilarity is always less than its base dissimilarity.

\begin{proposition}
Let $L<T$ be non-negative integers, and $0 \in E \subset \mathbb{R}$. Then
\beas
d_{L,E}(X,Y) \leq d(X,Y)
\enas
where $d_{L,E}(\cdot,\cdot)$ and $d(\cdot,\cdot)$ are defined as in \eqref{4TaStiC} and \eqref{basedist}, respectively.
\end{proposition}

\proof
Since $0 \in E$, the minimum in \eqref{4TaStiC} contains the term $d\left(X_{(0,0)},Y_{(0,0)}\right) = d(X,Y)$. This completes the proof. 

\bbox

The following shows that the base dissimilarity used in this work is not a mathematical distance. 
\begin{proposition}
The base dissimilarity as defined in \eqref{basedist} is a mathematical distance if and only if $\alpha=0$.
\end{proposition}
\proof
It is clear that when $\alpha=0$, $d(\cdot,\cdot)$ reduces to the weighted Euclidean distance. When $\alpha=1$, $d_{\Corr}((a,0,0),(b,0,0)) = 0$ for any nonzero $a \ne b$. This violates the second property in the definition of distance.
For $0<\alpha \le 1$, the triangle inequality is not satisfied. For instance, we let $X = (a,0,0)$, $Y = (a,a,0)$, and $Z = (0,a,0)$ for some $a>0$. Then $d_{\Corr}(X,Y) = 0.5 = d_{\Corr}(Y,Z) = 0.5$, and $d_{\Corr}(X,Z) = 1.5$. Hence, $d_{\Corr}(X,Z) - d_{\Corr}(X,Y) - d_{\Corr}(Y,Z) = 0.5 > 0$, which violates the triangle inequality for $\alpha=1$. It is also clear to see that $d_{weu}(X,Y) + d_{weu}(Y,Z) - d_{weu}(X,Z)$ converges to zero as $a \rightarrow 0$. Hence, for any $0<\alpha<1$, we can always find $a$ such that \\ $\alpha d_{\Corr}(X,Z) + (1-\alpha)d_{weu}(X,Z) > \alpha d_{\Corr}(X,Y) + (1-\alpha)d_{weu}(X,Y) + \alpha d_{\Corr}(Y,Z) + (1-\alpha)d_{weu}(Y,Z)$.

\bbox

The next property confirms that $D$ obtained from Algorithm \ref{alg:alg1} at least has zero diagonal. The proof is clear, since $\Corr(X,X) = 1$ for all nonzero $X \in \mathbb{R}^T$.

\begin{proposition}
The 4TaStiC dissimilarity as defined in \eqref{4TaStiC} satisfies
\beas
d_{L,E}(X,X) = 0 \text{ \ \ for any  \ \ } X \in \mathbb{R}^T.
\enas
\end{proposition}

Finally, we provide an example showing that adding the trend traveling affects the final dendrogram of the hierarchical clustering and, therefore, the final clusters.

\begin{example}

Consider a dataset as in Table \ref{tab:hclsutex} and plotted in Figure \ref{fig:hclust}. Each data point is generated from a form
\beas
y = at + b,
\enas
where $t=1,2,3,4,5$. $a = 0.2$, $0.1$, and $0.6$ for the first two rows, the middle row, and the last two rows, respectively. $b$ for the first two rows is generated from a multivariate normal with a mean of 7, a variance of 0.6, and a correlation of $5/6$. $b$ for the last three rows is generated from a multivariate normal with a mean of 7, a variance of 0.8, and a correlation of $3/4$. As we intend to classify data points by patterns, one possible label is $(1,1,2,2,2)$. Without the trend traveling, the 4TaStiC yields a clustering label $(1,1,1,2,2)$. With the trend traveling parameter $E = \{-0.4,0,0.4 \}$, the 4TaStiC gives the expected label.

\begin{figure}[h!]
  \centering
  \begin{minipage}[b]{.45\linewidth}
    \centering
    \begin{tabular}{cccccc}
      \hline\hline
       & T1 & T2 & T3 & T4 & T5 \\
      \hline
      $x_1$ & 7.59 & 7.72 & 6.27 & 6.07 & 8.51 \\
      $x_2$ & 7.78 & 7.76 & 6.93 & 6.04 & 8.37 \\
      $x_3$ & 7.63 & 7.79 & 7.39 & 6.58 & 5.79 \\
      $x_4$ & 7.96 & 8.65 & 9.10 & 9.42 & 8.25 \\
      $x_5$ & 8.18 & 9.19 & 9.01 & 9.47 & 9.20 \\
      \hline\hline
    \end{tabular}
    \captionof{table}{A dataset of example 1}
    \label{tab:hclsutex}
  \end{minipage}%
  \hfill
  \begin{minipage}[b]{.45\linewidth}
    \centering
    \includegraphics[width=\linewidth]{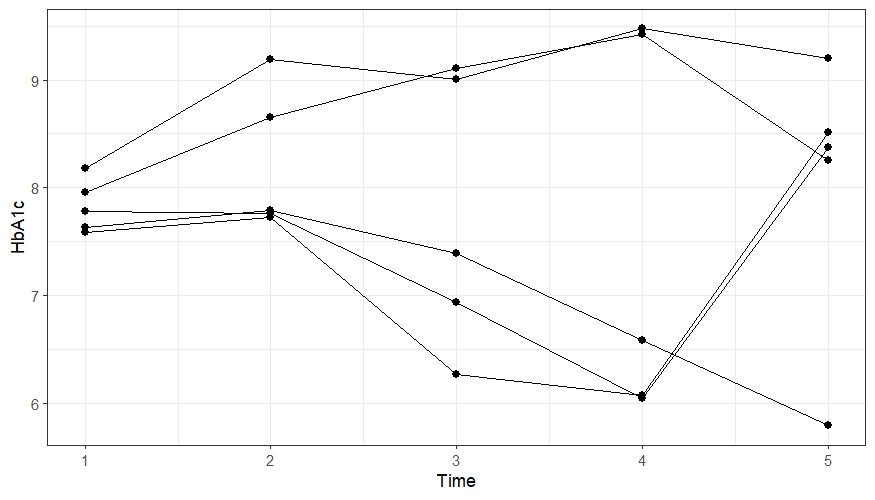}
    \captionof{figure}{The plotted data of example 1.}
    \label{fig:hclust}
  \end{minipage}
\end{figure}

\end{example}

\section{Experimental results} \label{sec:exp}

\begin{figure*}[htp]
\centering
\resizebox{0.7\textwidth}{!}{%
\begin{tabular}{ccc}
\hline\hline
Class1 & Class2 & Class3 \\
\hline 
\includegraphics[width=4cm]{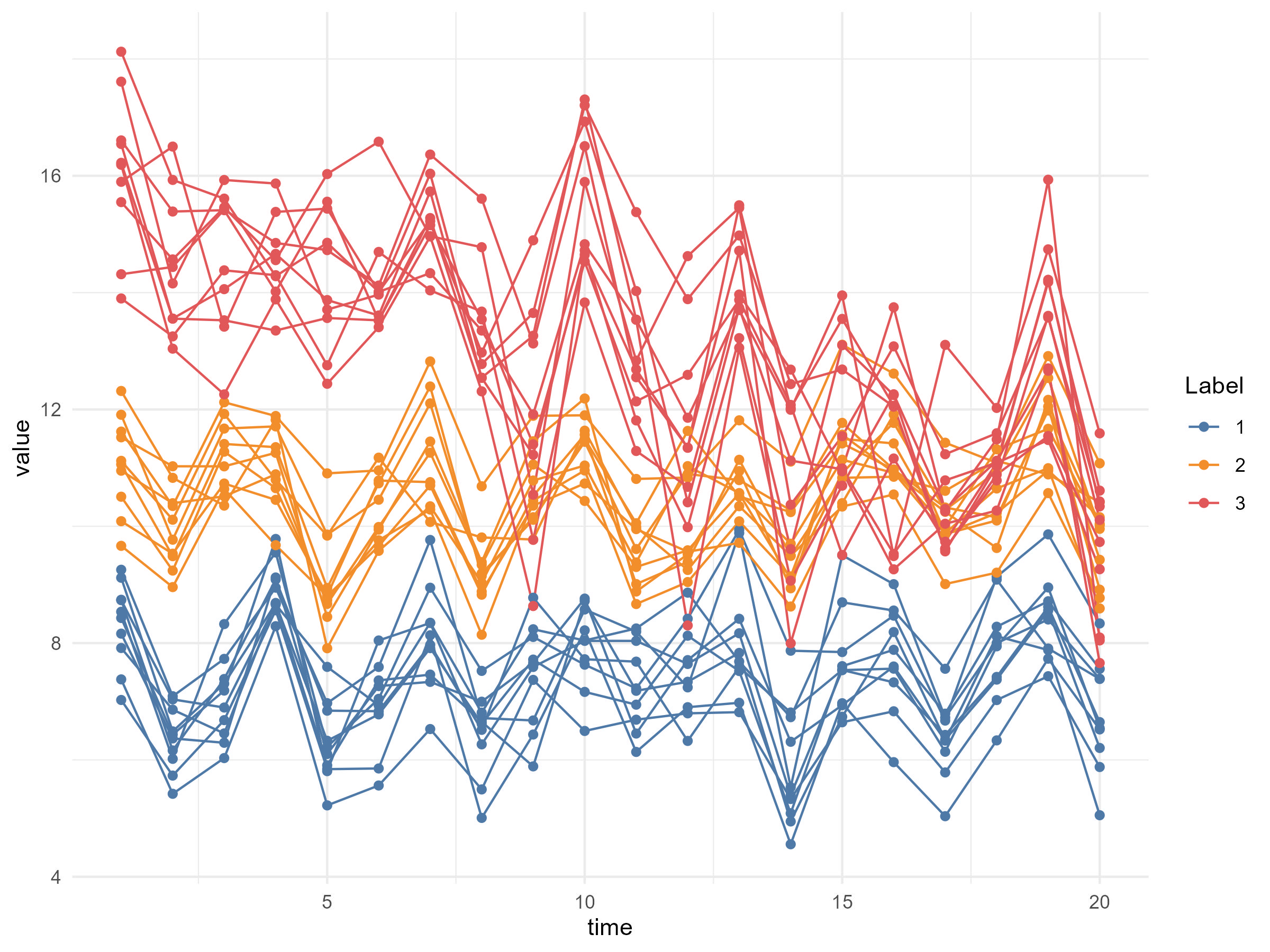} &\includegraphics[width=4cm]{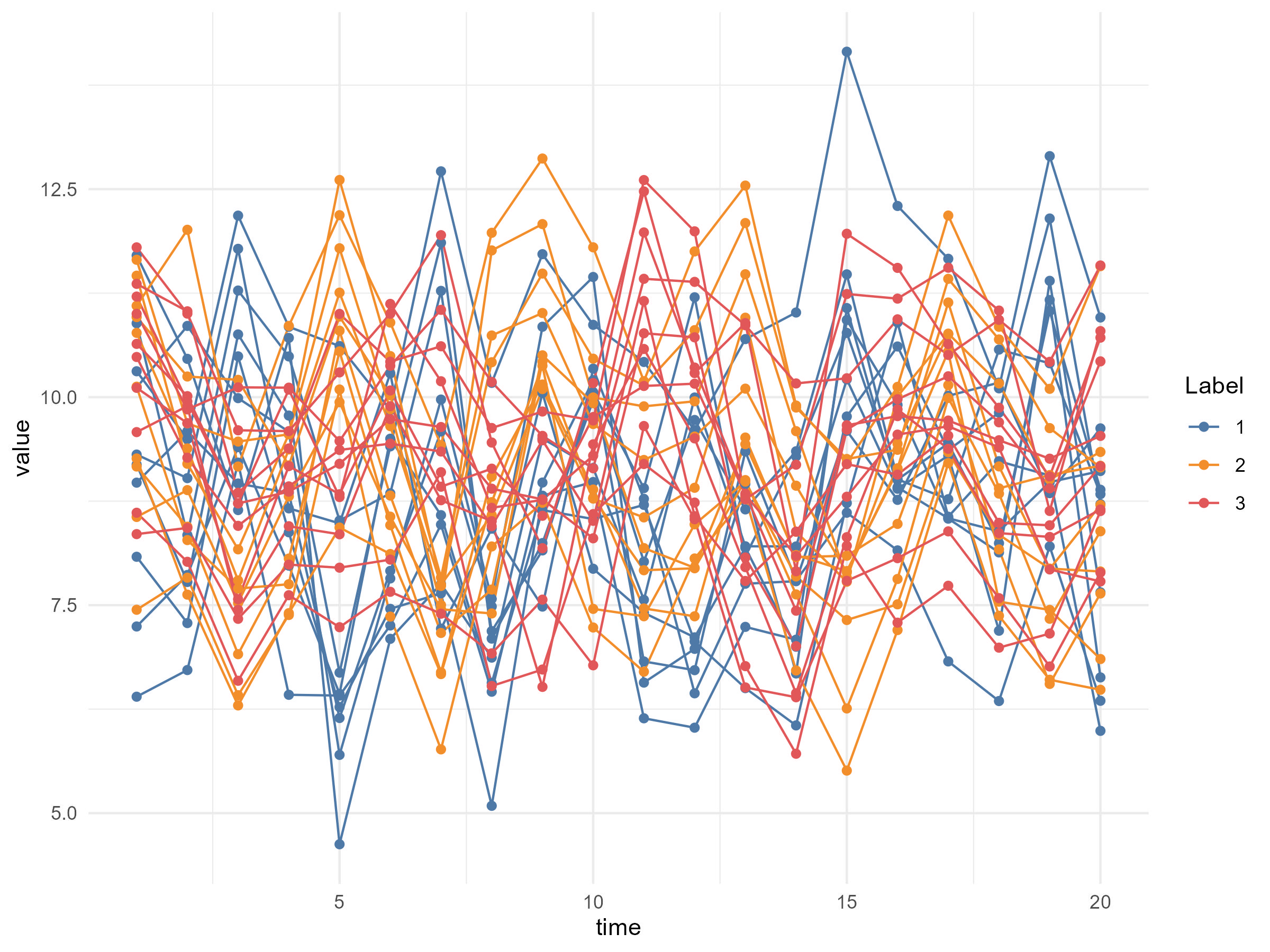}  & \includegraphics[width=4cm]{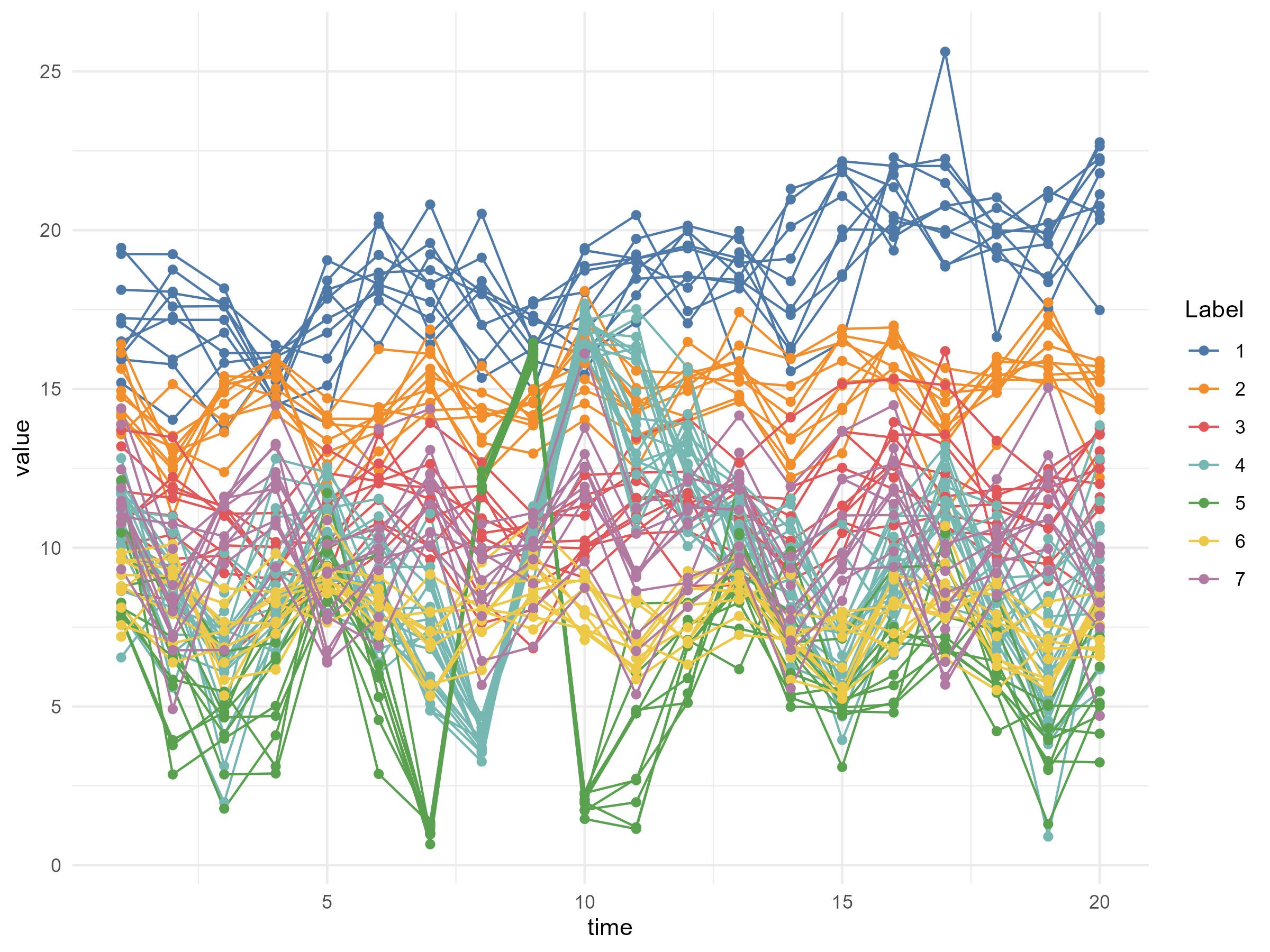}\\
 OG1\_1&OG2\_1  &OG3\_1 \\
\includegraphics[width=4cm]{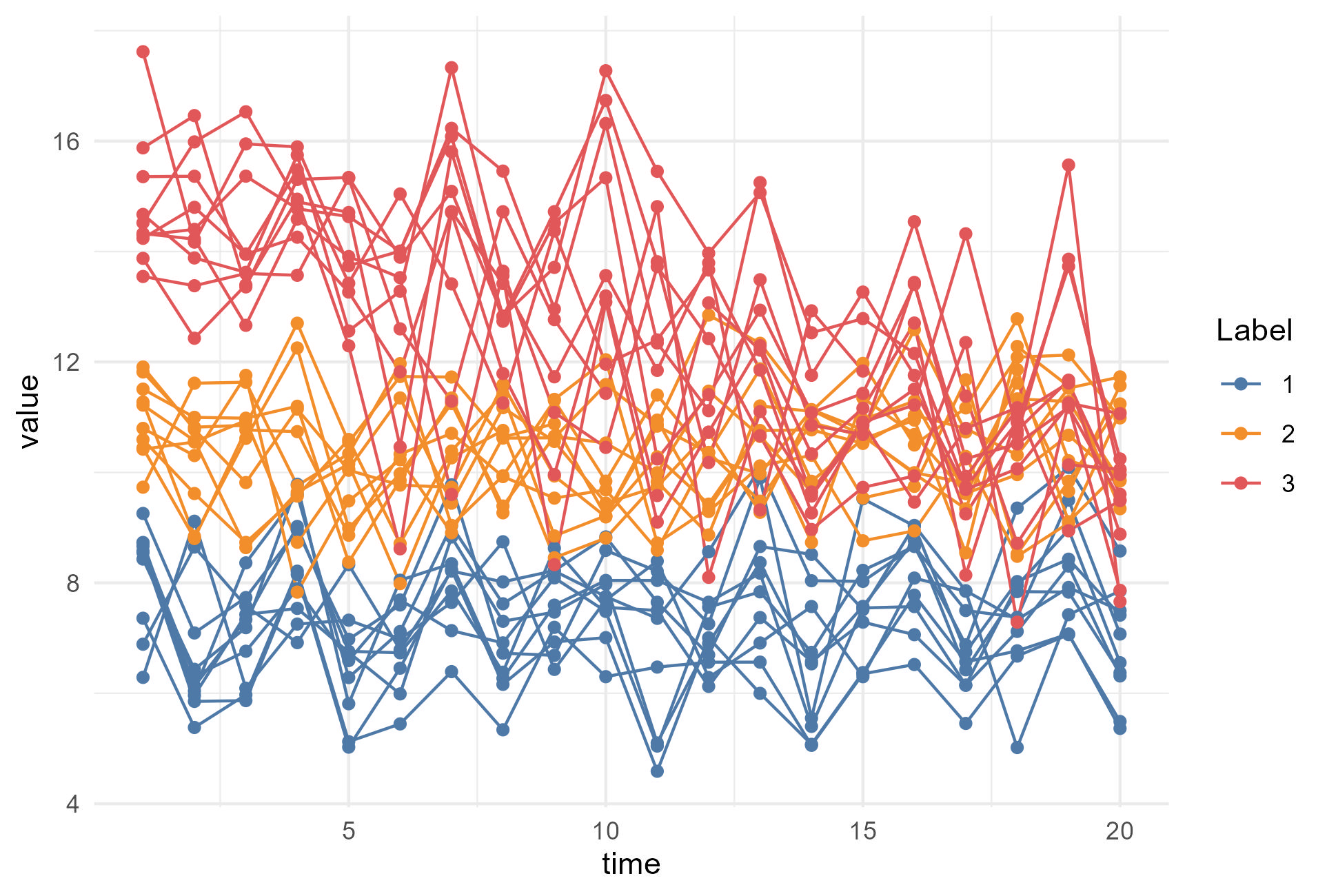} &\includegraphics[width=4cm]{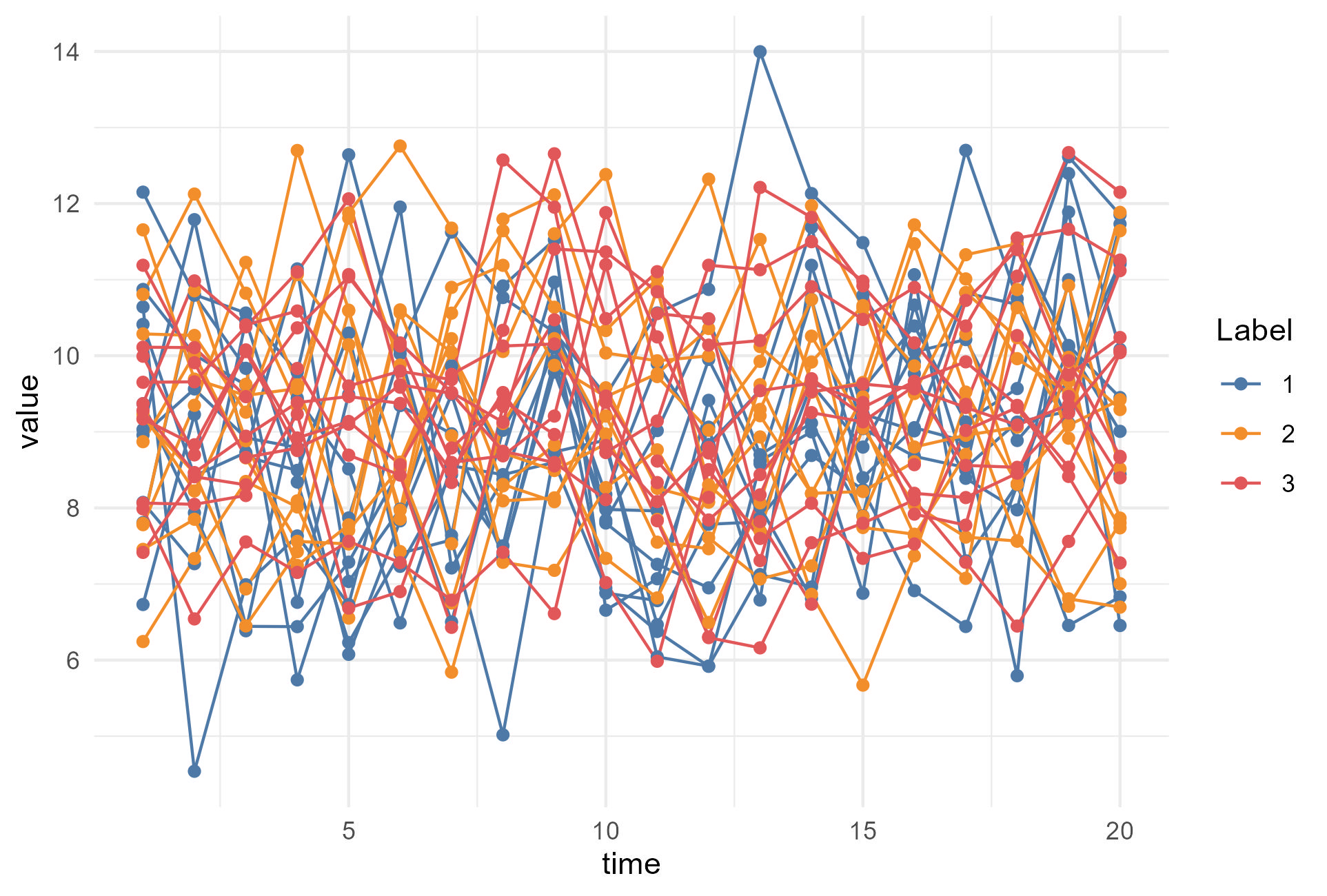}  & \includegraphics[width=4cm]{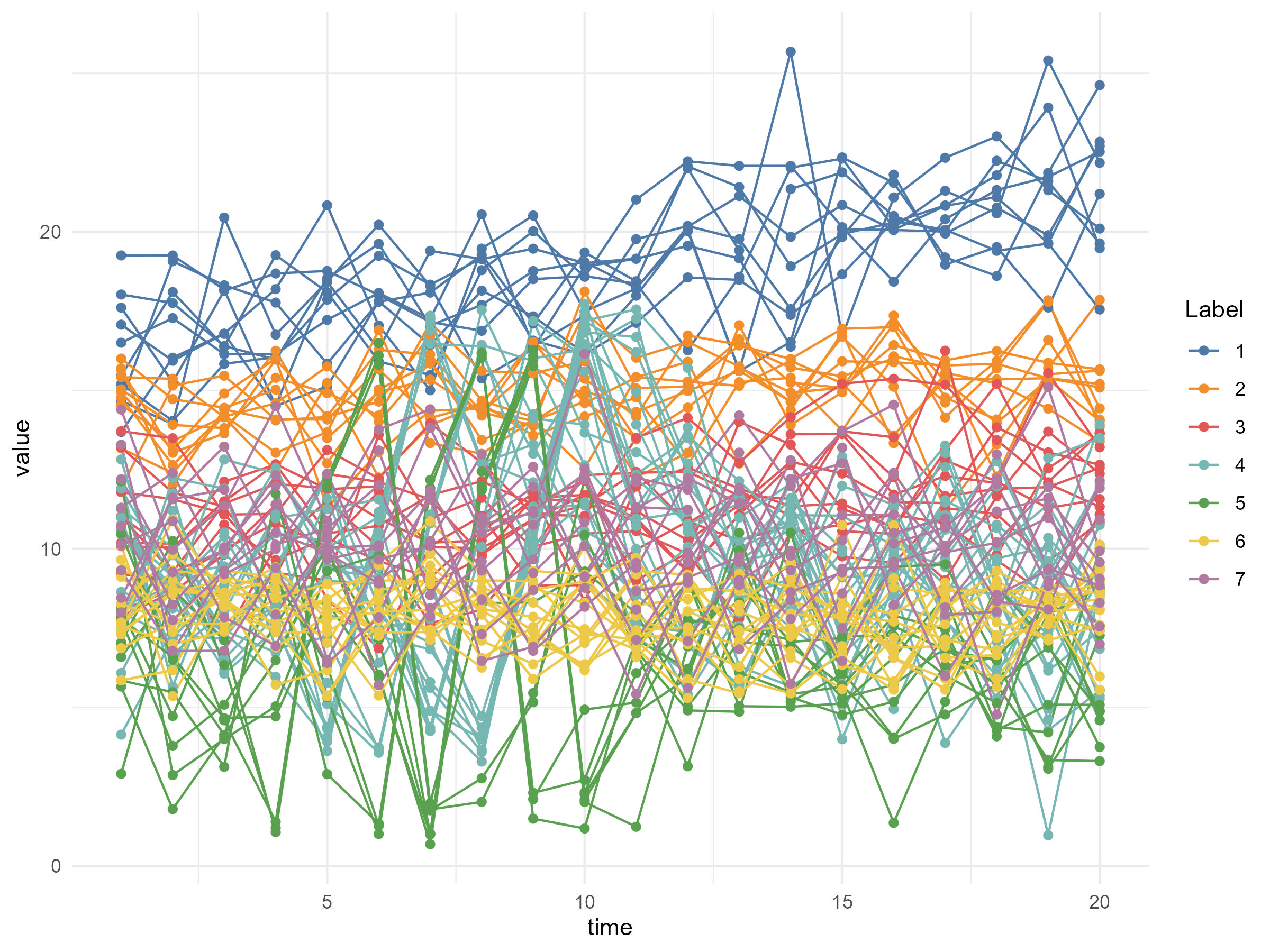}\\
 G1\_1&G2\_1  &G3\_1 \\
 \text{$ n = 30, \hspace{0.1cm} C= 3$}&\text{$ n = 30, \hspace{0.1cm} C= 3$ } & \text{$ n = 90, \hspace{0.1cm} C= 7$} \\
 \includegraphics[width=4cm]{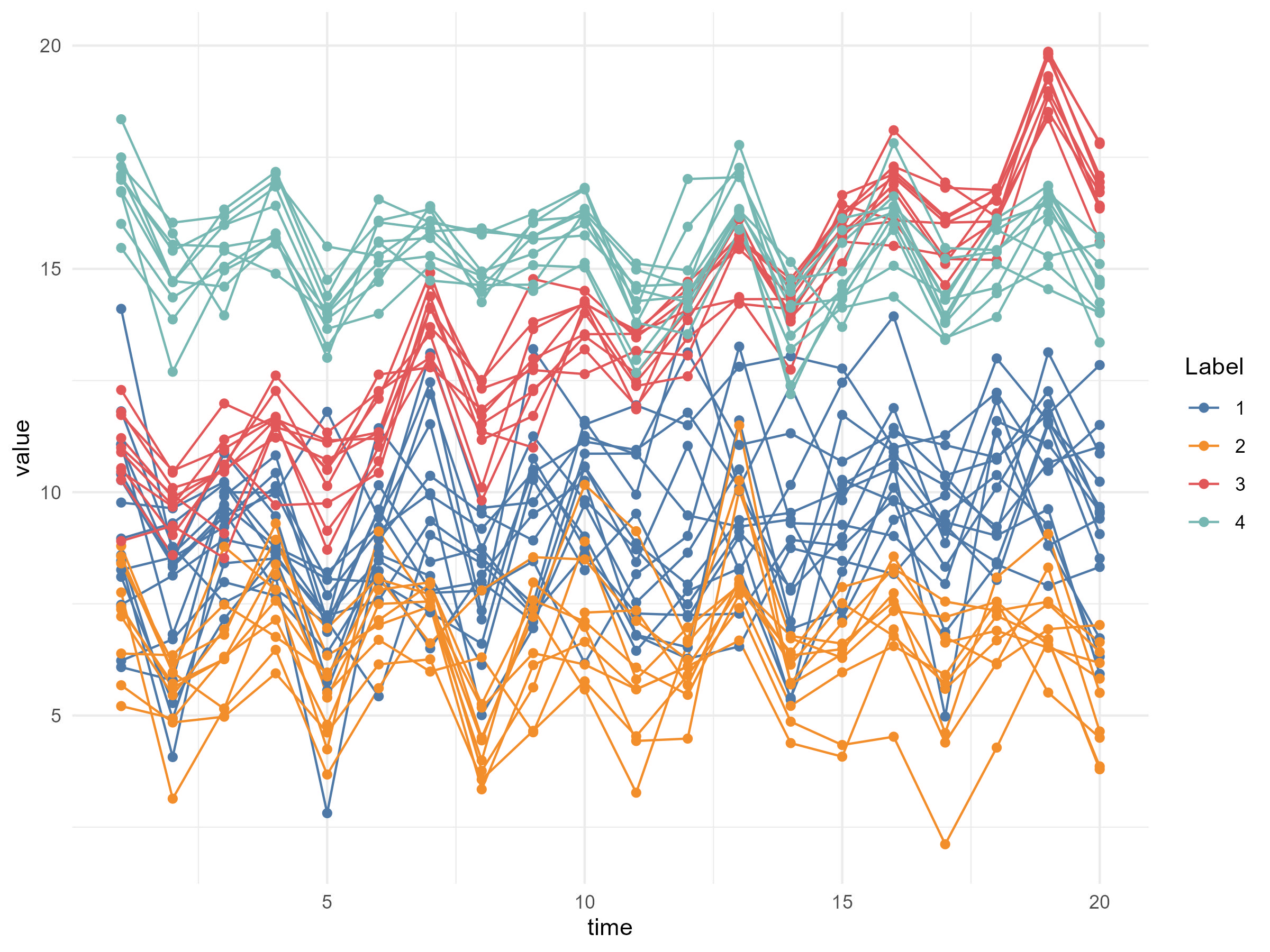} &\includegraphics[width=4cm]{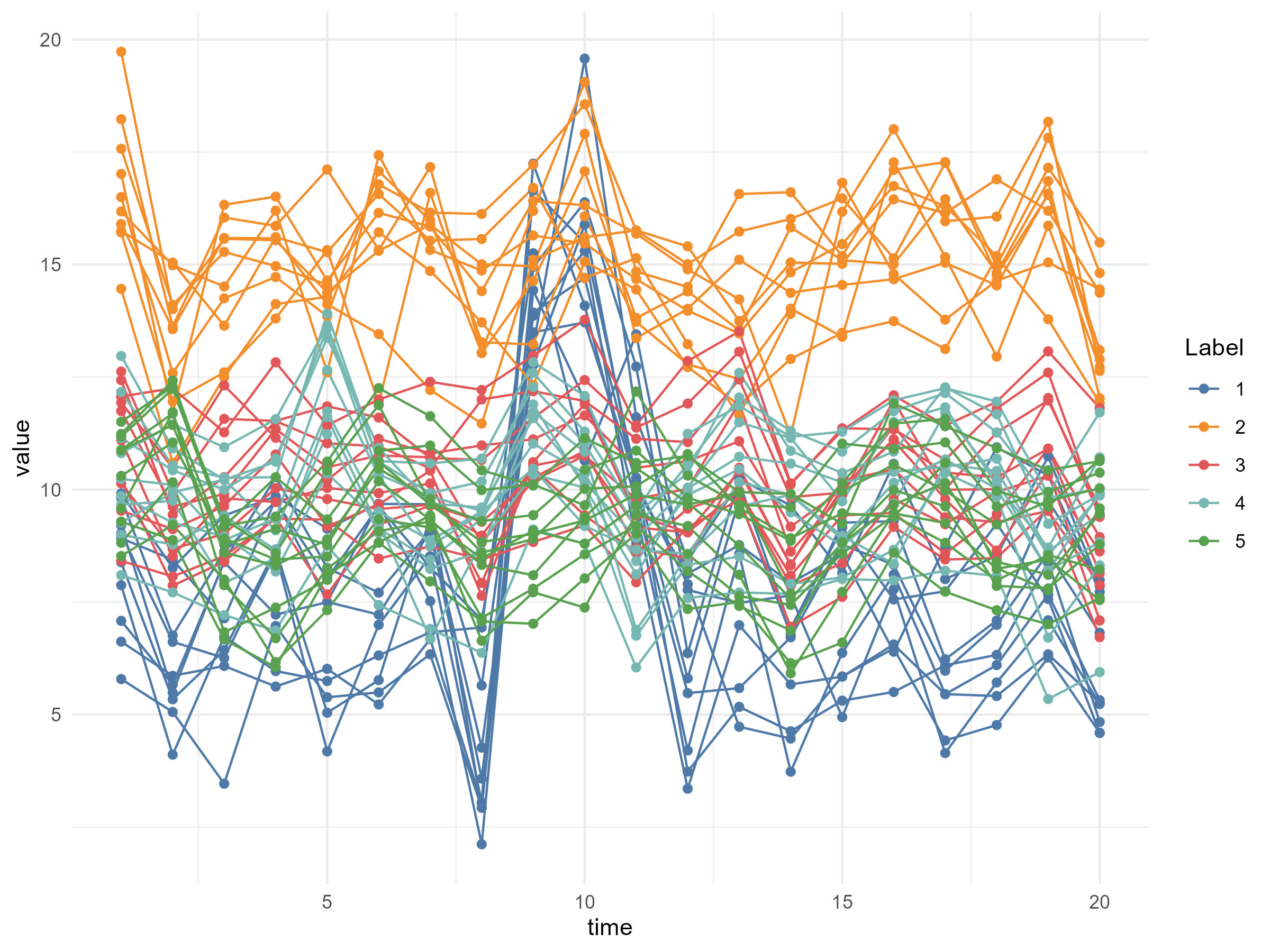}  & \includegraphics[width=4cm]{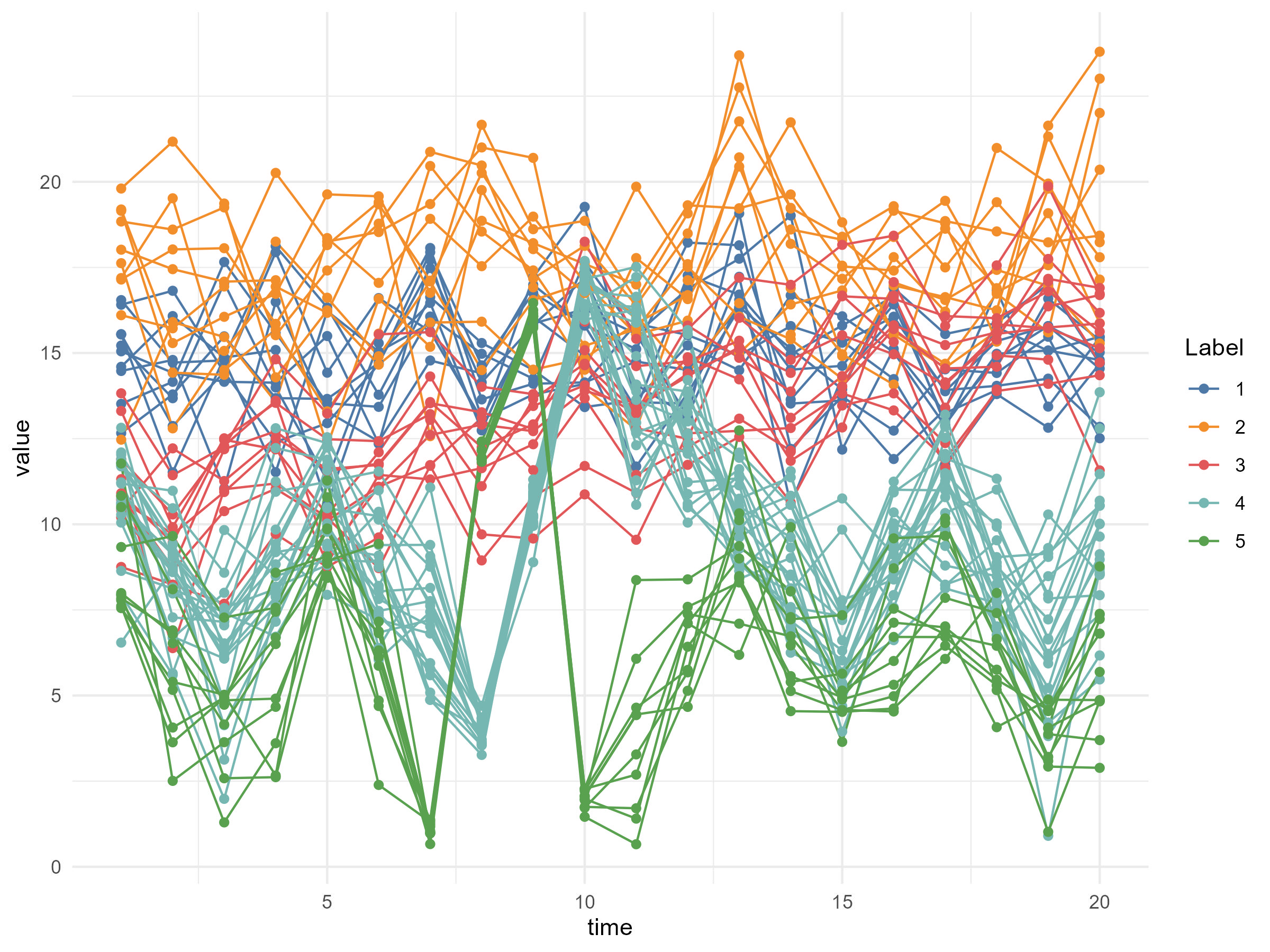}\\
 OG1\_2&OG2\_2  &OG3\_2 \\
\includegraphics[width=4cm]{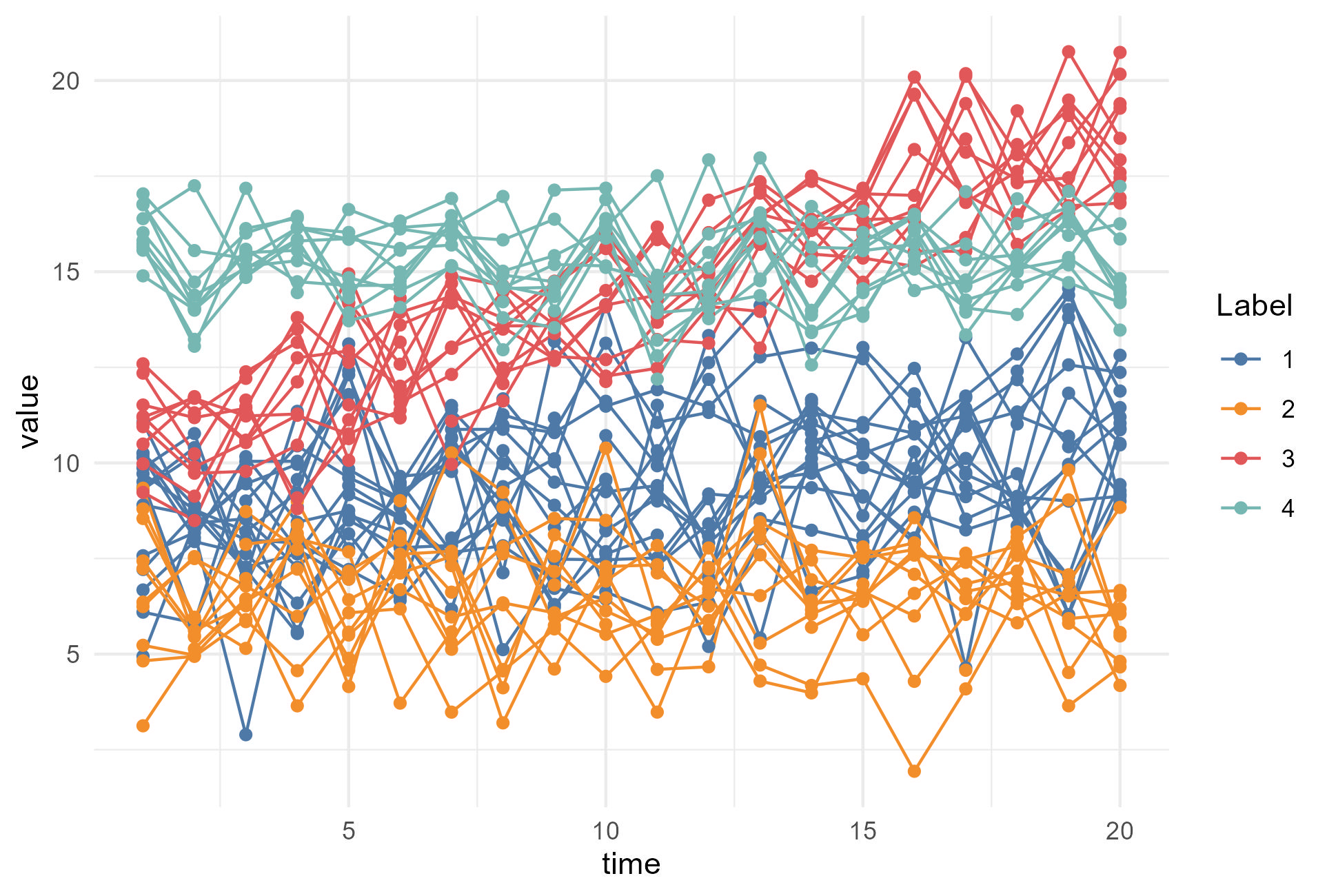} &\includegraphics[width=4cm]{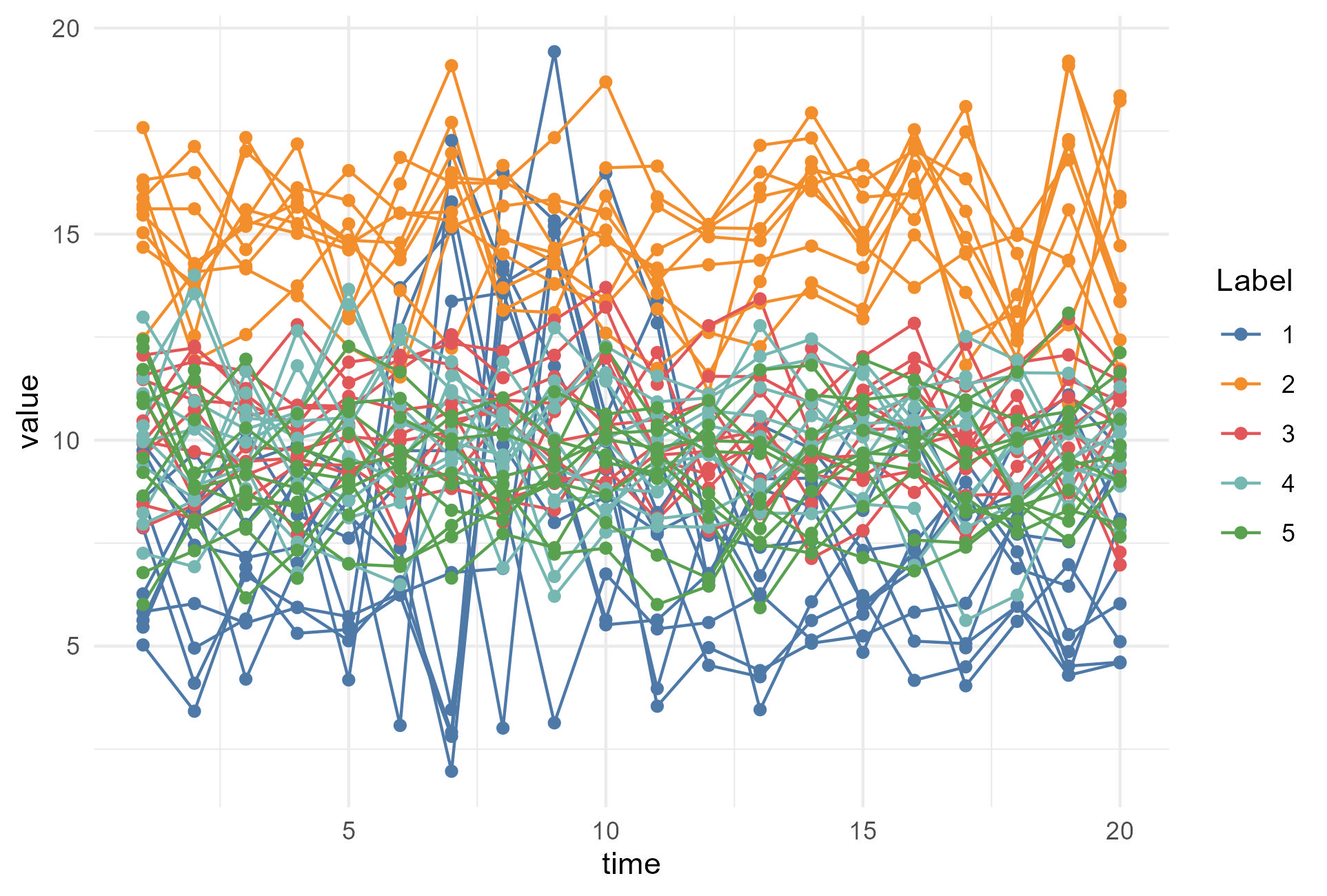}  & \includegraphics[width=4cm]{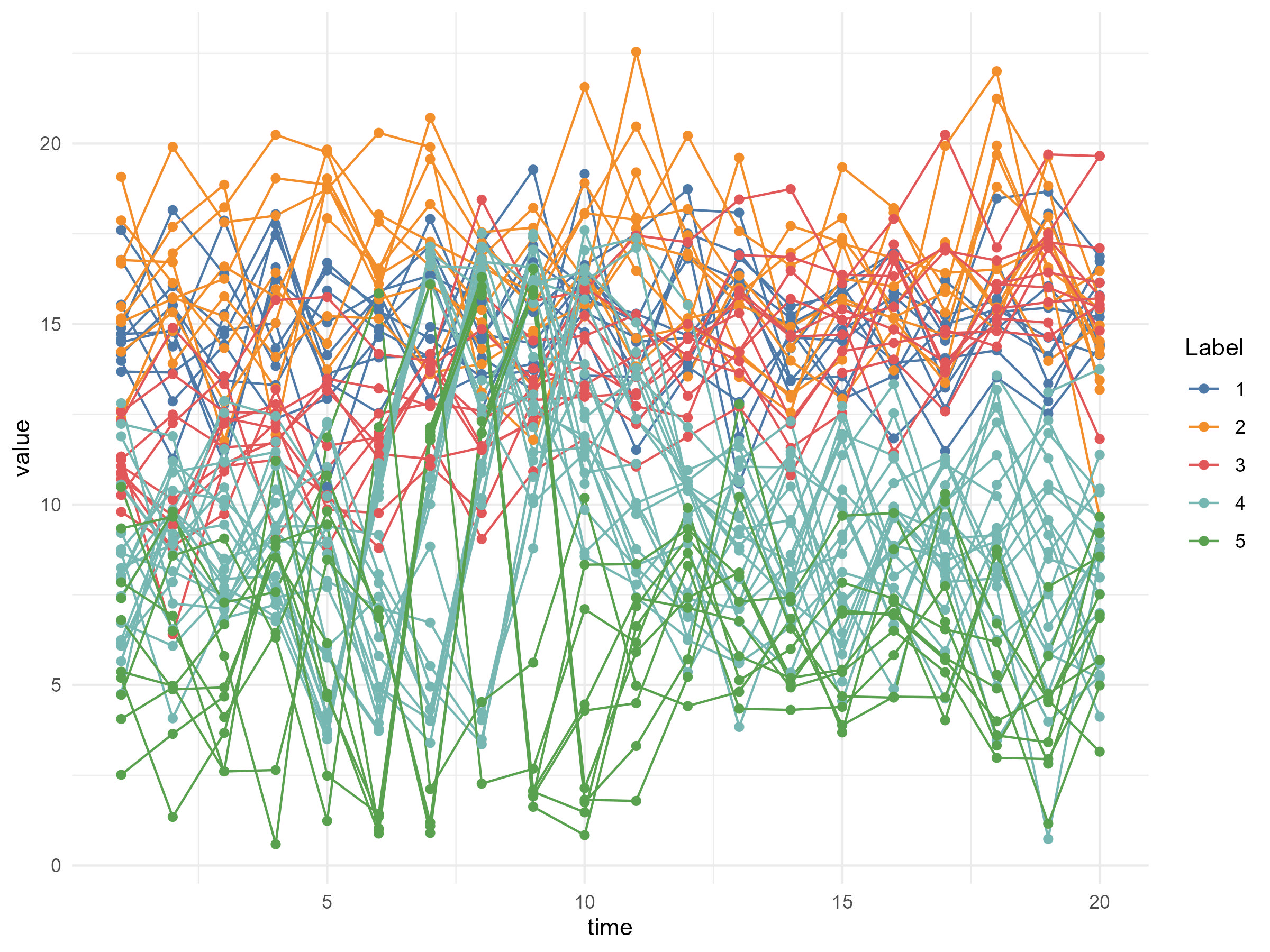}\\
 G1\_2&G2\_2  &G3\_2 \\
 \text{$ n = 45, \hspace{0.1cm} C= 4$}&\text{$ n = 50, \hspace{0.1cm} C= 5$ } & \text{$ n = 60, \hspace{0.1cm} C= 5$} \\
 \includegraphics[width=4cm]{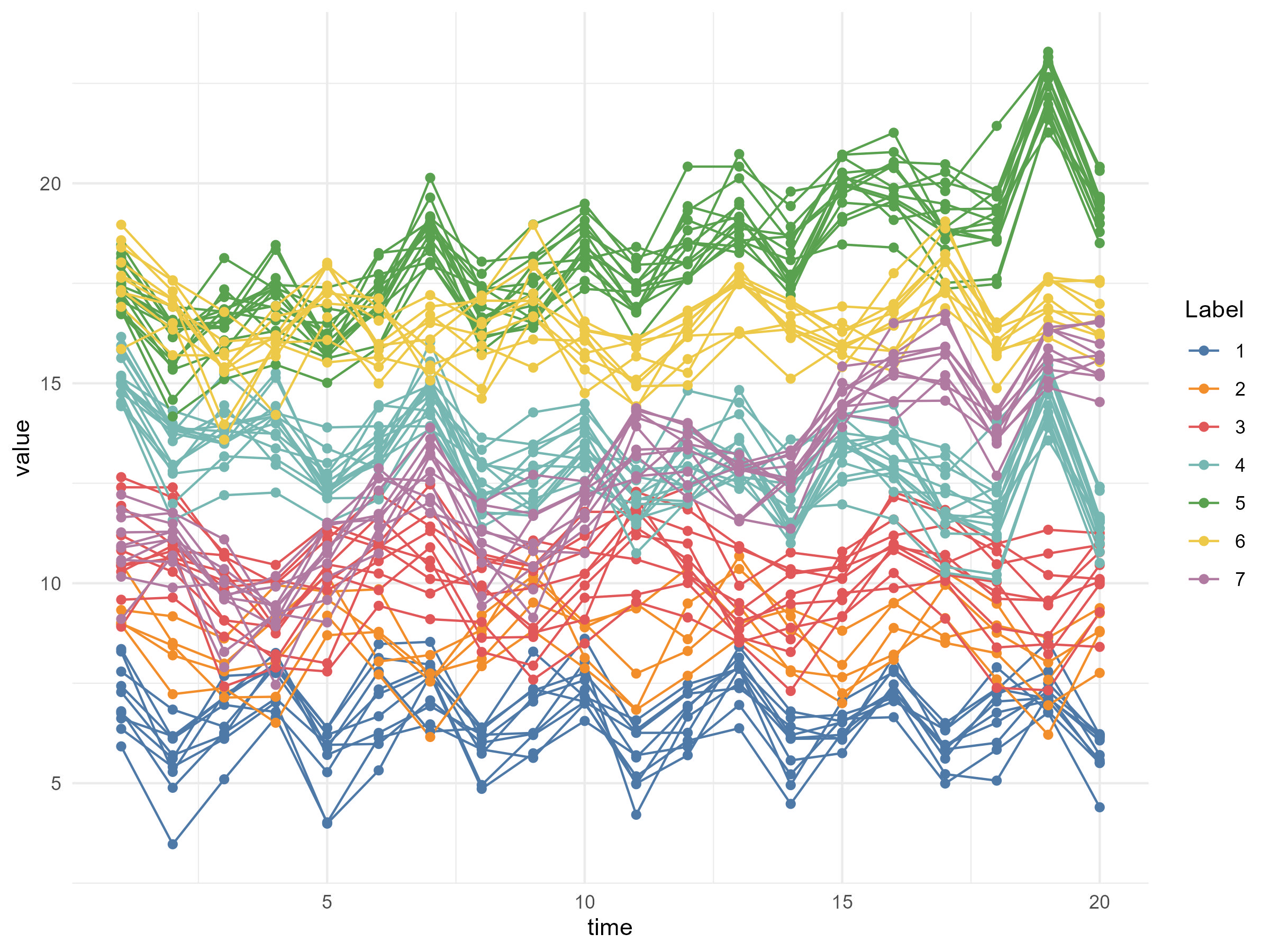} &\includegraphics[width=4cm]{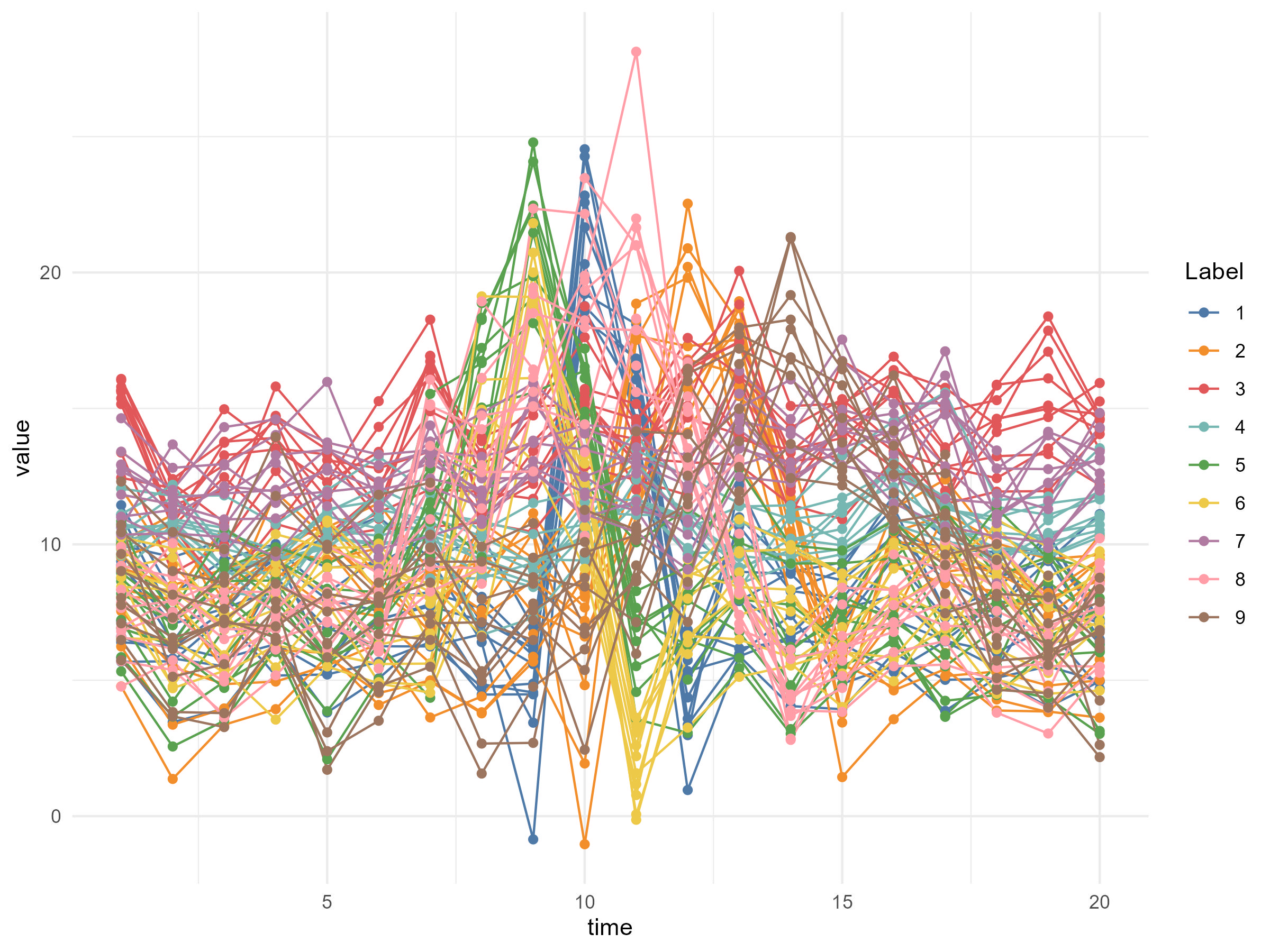}  & \includegraphics[width=4cm]{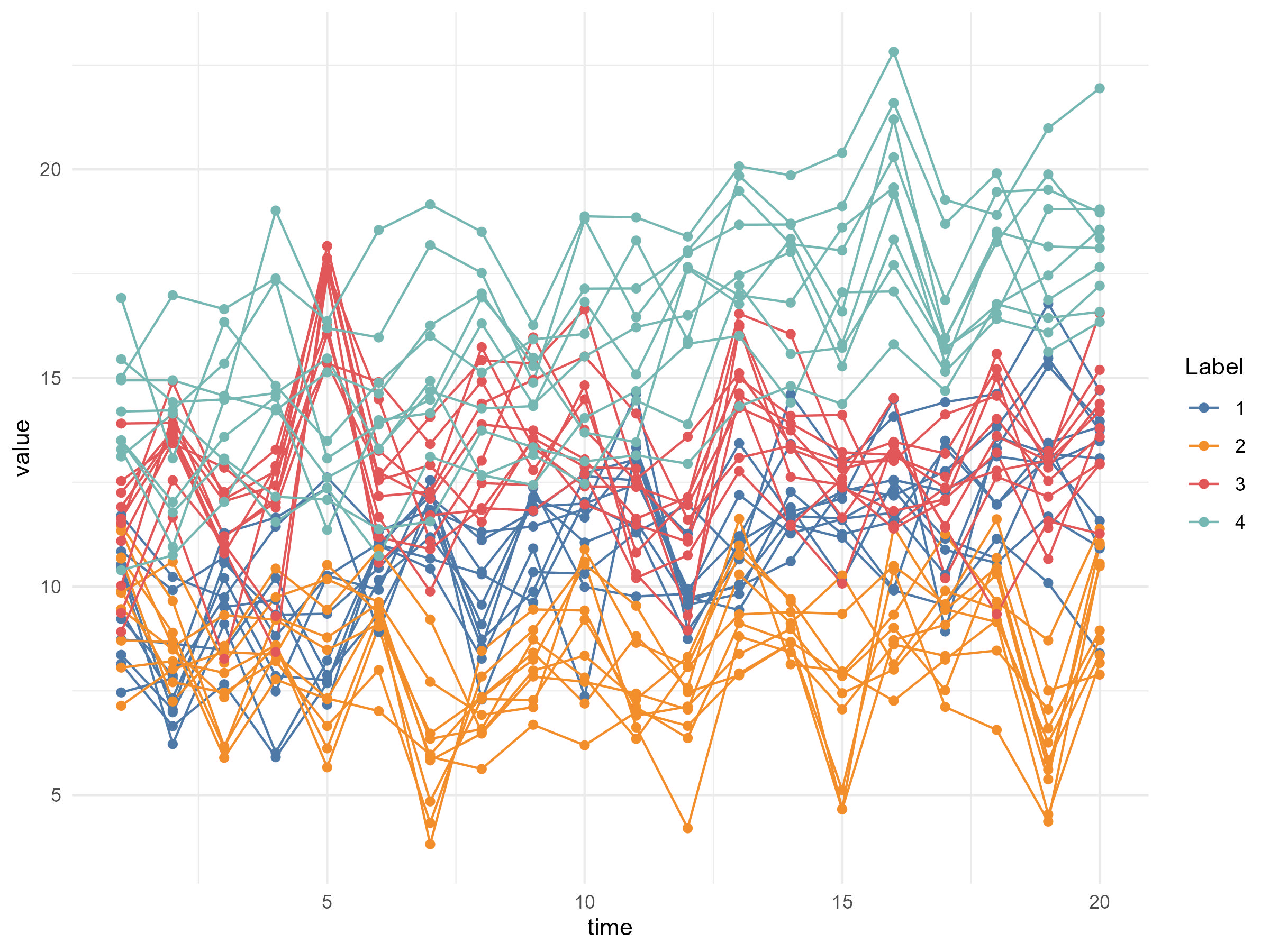}\\
 OG1\_3&OG2\_3  &OG3\_3 \\
 \includegraphics[width=4cm]{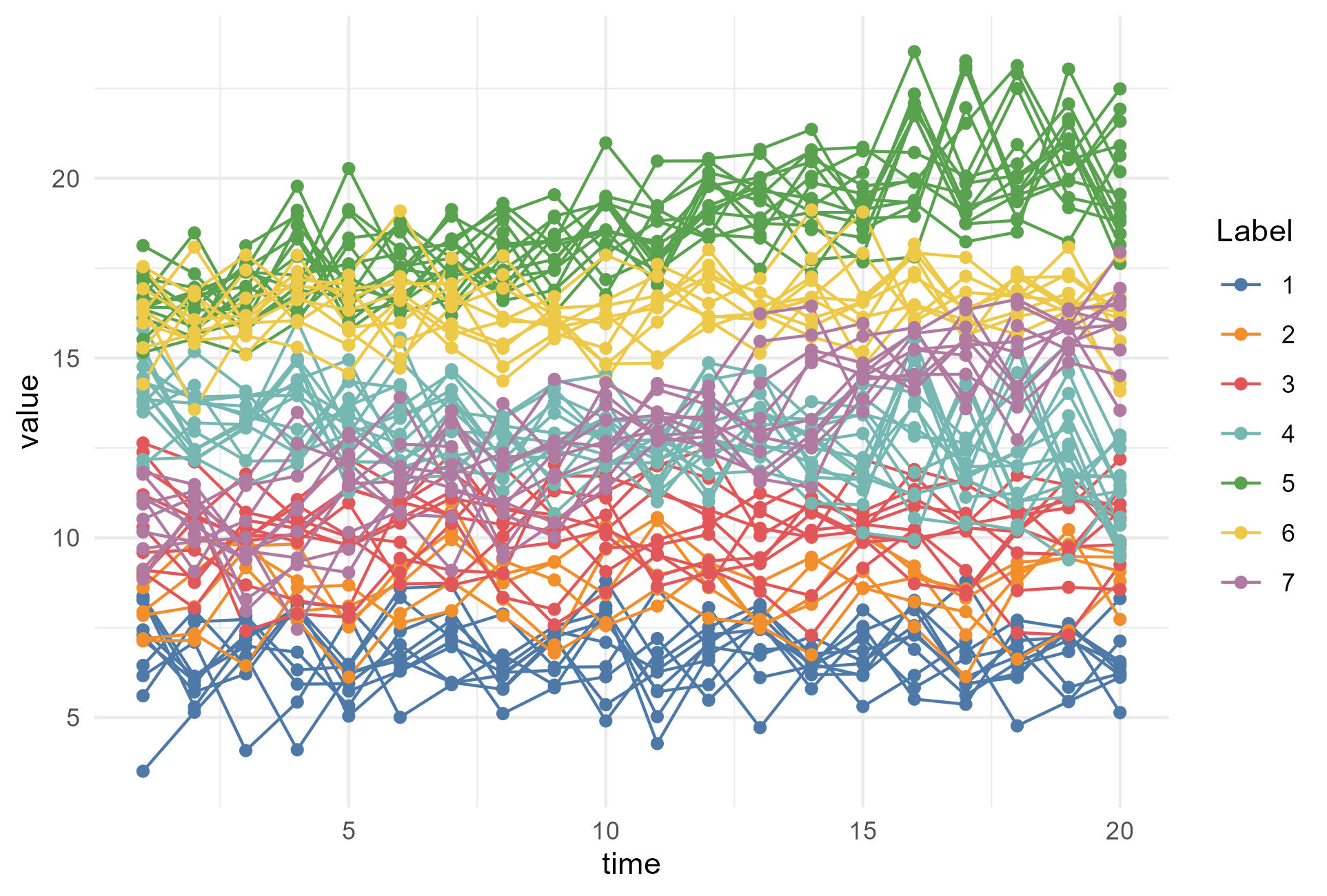} &\includegraphics[width=4cm]{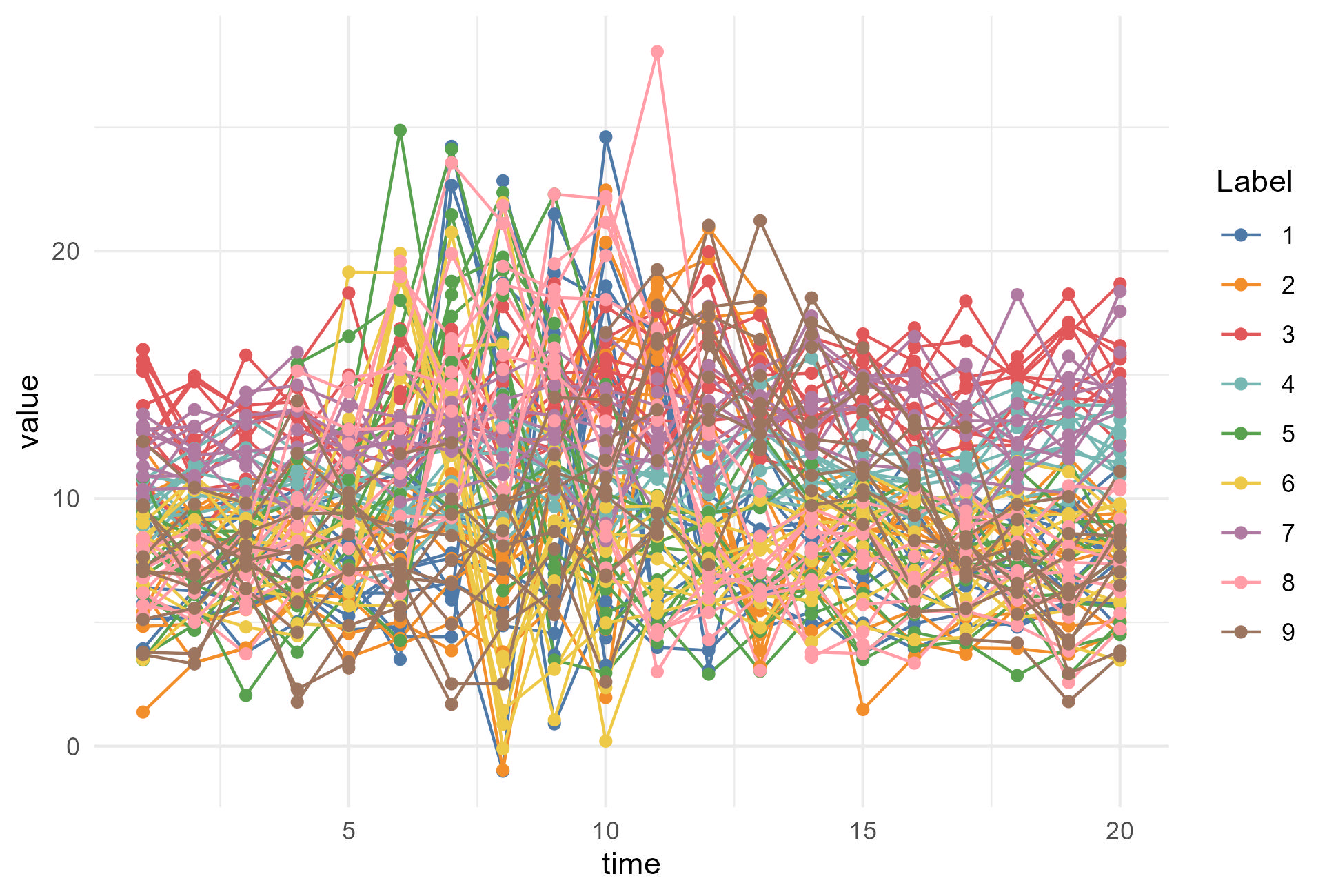}  & \includegraphics[width=4cm]{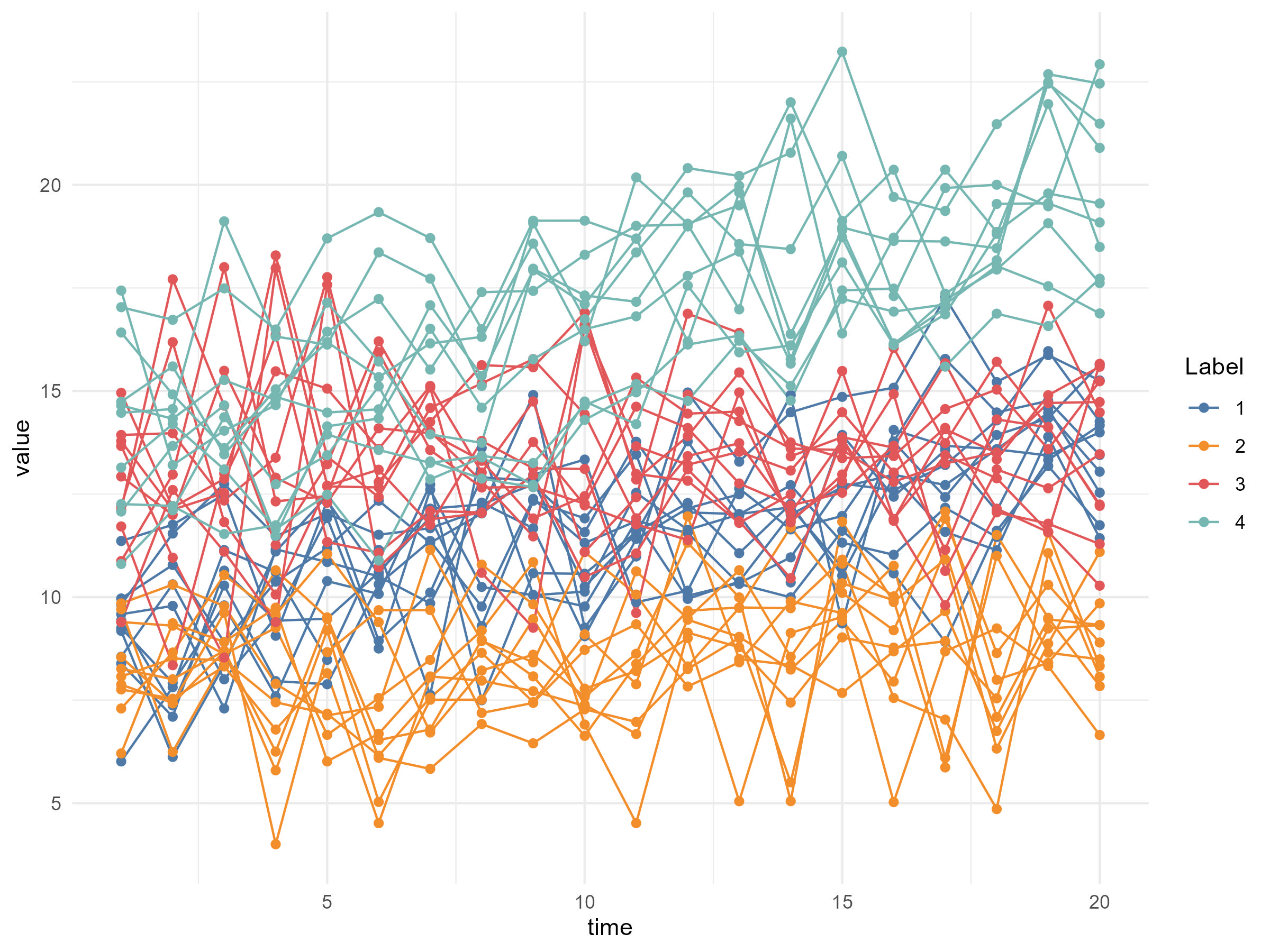}\\
 G1\_3&G2\_3  &G3\_3 \\
 \text{$ n = 75, \hspace{0.1cm} C= 7$}&\text{$ n = 90, \hspace{0.1cm} C= 9$ } & \text{$ n = 40, \hspace{0.1cm} C= 4$} \\
\hline\hline
\end{tabular}
}
\text{Note: $n$ and $C$ are the number of data points and the true number of clusters,}
\text{respectively. OG refers to the original image.}
\caption{Artificial datasets}
\label{fig:art_plots}
\end{figure*}

In this section, we evaluate 4TaStiC's performance on simulated artificial datasets, as shown in Figure \ref{fig:art_plots}. Each dataset is generated using a sine wave-based structure to represent periodic patterns (seasonality), while Gaussian noise and random variation introduce variability. Additionally, variations in means, standard deviations, seasonal effects, trends, and noise intensity create diverse data characteristics. The datasets are categorized into three distinct classes based on these factors, as follows.

\begin{list}{}{} \label{datagroup}
\item {\bf Class 1 (Well-separated data by Euclidean distance): } This group emphasizes clear cluster separation based on Euclidean distances, primarily achieved through distinct mean-level differences.

\item{\bf Class 2 (Well-separated data by correlation): } This group introduces increased complexity by incorporating correlated structures, diverse seasonal patterns, varying noise intensities, and up–down patterns.

\item{\bf Class 3 (Complicated data involving patterns, trends, and peaks.): } This group is generated by combining the complications from the previous two groups, illustrating a combination of seasonal effects, trends, and distinct peaks, leading to dynamic variations in datasets. 
\end{list}
Class 3 is intentionally generated to imitate patients' laboratory results, which sometimes exhibit similar patterns, trends, and peaks to those of close laboratory levels. Tiny trends and noises are added to the data to mimic real scenarios where every single pair of patients differs at least slightly. All the sample time series are randomly shifted so that even if two patients exhibit the same patterns and result values, the visit times are different, as no two patients always visit the doctor at the same time. Figure \ref{fig:ex1} illustrates this idea .

Although these datasets are generated to imitate those including real patients, each cluster exhibits clearer characteristics than real data—these are the characteristics we intend to subject to unsupervised classification. The purpose this is to demonstrate that our proposed algorithm can correctly classify patients according to our intuitive understanding. We can then be confident in the validity of the classification of diabetes patients described in the application section.

\subsection{Sensitivity analysis} \label{sen_test}

We first perform a sensitivity analysis to observe how small changes in weight $\alpha$ and the tilting parameter $\epsilon$ affect the clustering performance in terms of accuracy and Adjusted Rand Index (ARI). This analysis also examines the effect of adding penalties to the trend traveling term. We apply the default $\alpha$ formula as in \eqref{alpha} with $p \in \{0, 0.01,0.02,\ldots,0.10 \}$, and let $L=3$, $E=\{-\epsilon,0,\epsilon \}$ with $\epsilon \in \{0,0.025,0.05,0.075,0.1 \}$, and $C\in \{0,0.5,1 \}$. We maintain a small value of $p$ since a larger value reduces the importance of the correlation term as in \eqref{alpha}, and we intend to detect patterns using the correlation. We also note that the set parameter $E$ should be selected based on the maximum acceptable distance between the two time series for them to be assigned to the same cluster, given that they exhibit similar patterns and trends, etc. For instance, two patients whose difference in HbA1c is greater than 1.5 should not be considered the same. With 12 time steps and as in \eqref{TrendT}, this rotates the last time point by $11 \times \epsilon \in \{0,...,1.1 \}$. Regardless of pattern similarity, rotating more than this should result in the patients not being matched or assigned to the same group. 

Table \ref{tab:sensitivitytable} presents the accuracy and  ARI of 4TaStiC for the 45 combinations of parameters on the three datasets in Class 3. We test the sensitivity of Class 3 since it is the most complicated, and we mainly focus on the performance in this class, as mentioned above. 4TaStiC has high sensitivity for a small $p < 0.05$ because it relies only on the correlation-based dissimilarity. However, the results show that 4TaStiC is quite stable for a larger $p$ value, with an accuracy varying within the range of about 0.05.   

The parameters $p=0.1$, $\epsilon=0.025$, and $C=0$ yield the highest accuracy and the best ARI on both G3\_1 and G3\_3, but does not perform well on G3\_2. The combination of parameters $p=0.09$, $\epsilon = 0.075$, and $C=0$ yields one of the most balanced performances overall across the three datasets. The results have low sensitivity around this set of parameters. For parameter $C$, we intentionally do not penalize the trend traveling because we aim to match two patients with the same pattern but slightly different trends. This explains why we used this combination of parameters for the diabetes patients' data.

Although we decide not to set $C>0$ to penalize the trend traveling in this work, we show that penalizing can help improve clustering results in some cases. Specifically, the italics in Table \ref{tab:sensitivitytable} show that $C>0$ yields a higher accuracy and ARI for some parameters. For instance, the parameters $p=0.05$, $\epsilon = 0.075$, $C \in \{0.5,1\}$ yield an accuracy of 0.97 and an ARI of 0.93 compared to the 0.83 and 0.77 achieved, respectively, with $C=0$ on G3\_1. This is useful in real situations where an excessive trend traveling parameter is accidentally used; in these cases, we protect against the effects of such an accident by assigning a penalty.

\begin{table}[H]
\centering
\includegraphics[width=16cm]{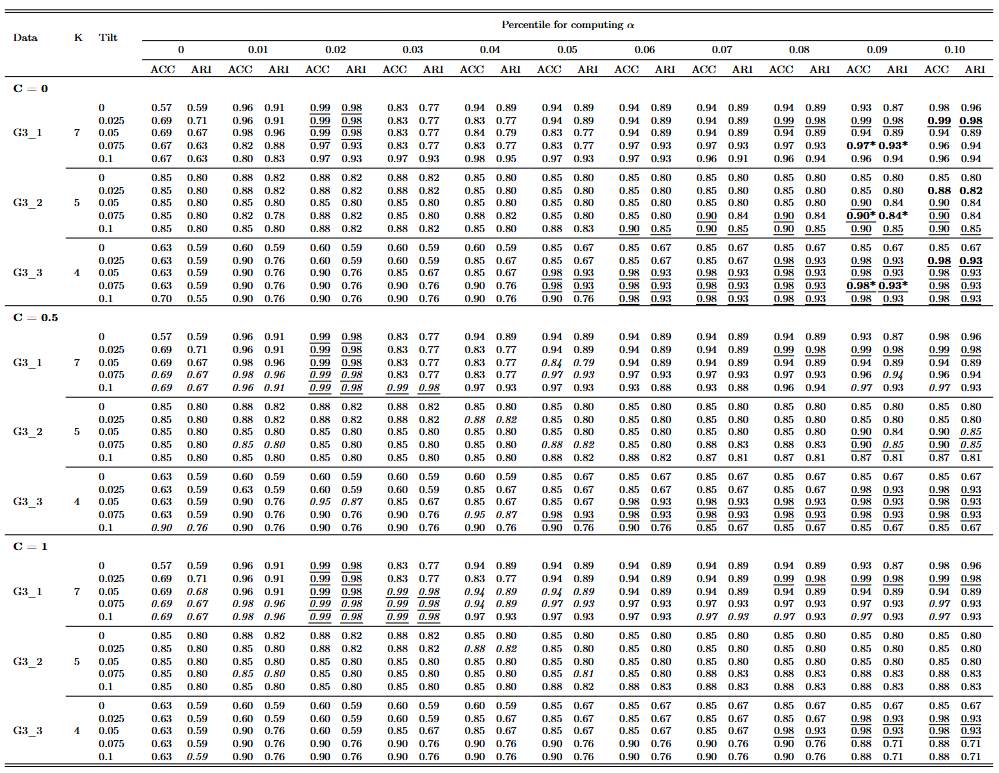}
\caption{Sensitivity analysis of the G3 datasets. The locations with the highest accuracy and ARI for each dataset are underlined. The parameters of interest are indicated in bold. The selected parameters are marked with stars. The places where penalizing yields better results are in italic type. }
\label{tab:sensitivitytable}
\end{table}

\subsection{Performance} \label{performance}
To confirm that our proposed method is suitable for clustering the data of diabetes patients based on our motivation, we test 4TaStiC's performance with the parameters $p=0.09$ and $E=\{-0.075,0,0.075\}$ on nine artificial datasets as detailed above. 

We test 4TaStiC's accuracy and ARI  by comparing them to the results of existing clustering methods, including K-means, hierarchical clustering with Euclidean distance, Pearson correlation dissimilarity, cross-correlation dissimilarity, DTW, GAK, and LPWC. The time traveling technique is added to all the hierarchical clustering algorithms with Euclidean, correlation-based, and 4TaStiC dissimilarities to allow for time series with similar patterns and generated values but different timelines to be assigned to the same group. It is not used for DTW, GAK, and LWPC, as the three dissimilarities incorporate different concepts of time warping. It does not apply to K-means, as centroids will no longer be computable once the time is shifted differently. The trend traveling concept is applied only to the hierarchical clustering algorithms with correlation-based dissimilarity and 4TaStiC, since it is unreasonable to tilt time series and compute a physical distance such as the Euclidean distance.

The results are presented in Table \ref{acctable} and show that 4TaStiC with both time and trend traveling achieves a superior performance on the datasets in Class 3, with accuracies of 0.97, 0.9, and 0.98, and ARI values of 0.93, 0.84, and 0.93, respectively. The physical distance-based algorithms, including K-means, hierarchical clustering with DTW, GAK, and the Euclidean distance, all perform well on the datasets in Class 1. However, 4TaStiC without trend traveling achieves the best performance across the three datasets in this class. 4TaStiC with $\alpha=1$ (pure correlation) certainly has the best overall performance on the datasets in Class 2. 4TaStiC with $\alpha<1$ also performs well for this class and even better than the pure correlation based on G2\_3. The higher accuracy and ARI make it clear that the time traveling technique is the key to grouping similar time series with unmatched time points. The trend traveling technique then often improves the precision slightly. 

Since the number of clusters is unknown in real applications, we perform the elbow method to illustrate that it can guide us towards the correct number. It can be observed from Tables \ref{acctable} and \ref{tbl:elbow artificial} that the elbow method often displays an elbow at the correct number of groups when the corresponding algorithm is accurate. As we focus more on the datasets in Class 3, the elbow method based on 4TaStiC with both time and trend traveling leads us to the correct number of groups for the last two datasets: G3\_2 and G3\_3. However, the elbow method based on the algorithm without trend traveling leads to the correct number for G3\_1. As there is no perfect way to detect the number of groups in cluster analysis, we rely on the elbow method and accept that it may sometimes lead to sub-optimal numbers of groups. We leave it to future work to discuss deeper methods, like cluster validity indices (CVIs), to detect the number of groups. We note here that existing CVIs do not apply directly to our case because of our use of the time and trend traveling technique (see \cite{NCI, BayesCVI} for further detail). This certainly affects how the CVIs measure inter-cluster and between-cluster distance and centroids.

\begin{table}[H]
\centering
\includegraphics[width=16cm]{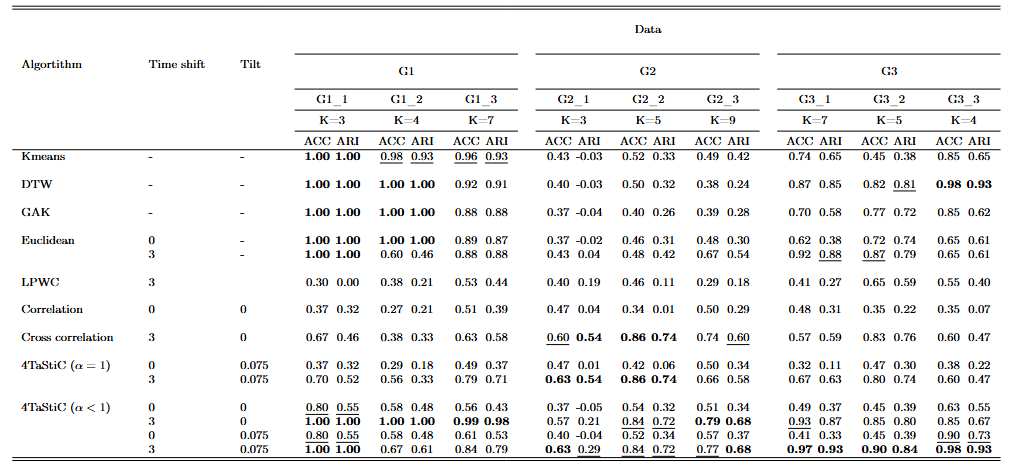}
\caption{Algorithms comparison. The highest and second-highest accuracy and ARI are bold and underlined, respectively.}
\label{acctable}
\end{table}

\begin{table}[H]
\centering
\includegraphics[width=15cm]{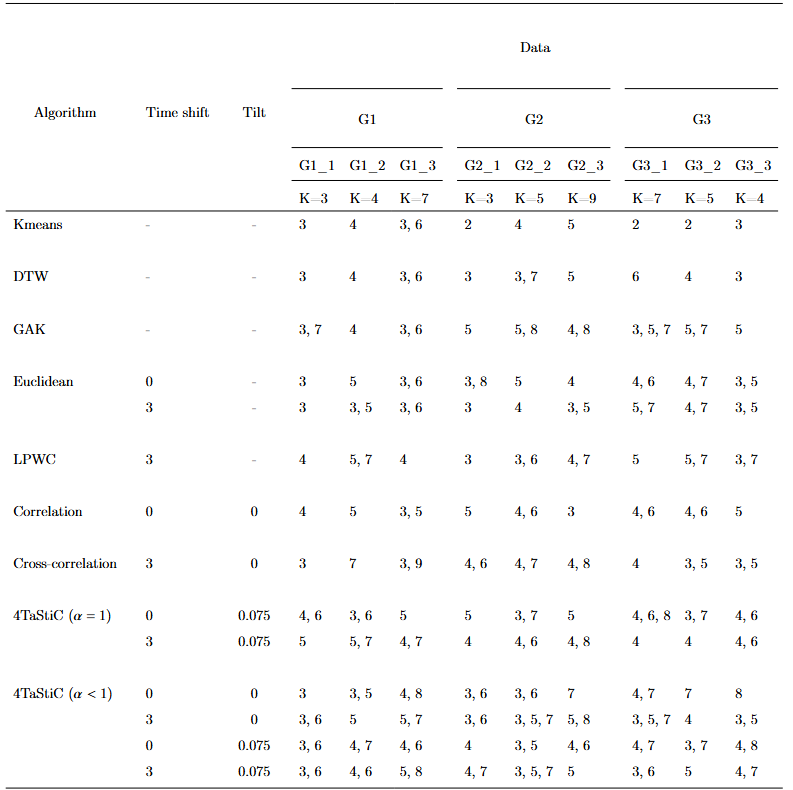}
\caption{Elbow method on artificial datasets. The three clearest elbow points (if applicable) are shown.}
\label{tbl:elbow artificial}
\end{table}

\section{Application to diabetes patients data} \label{sec:app}

In this section, we present an application to classify diabetes patients from Siriraj Hospital.

\subsection{Diabetes patients' data }

\begin{figure}[htp]
    \centering
    \includegraphics[width=13cm]{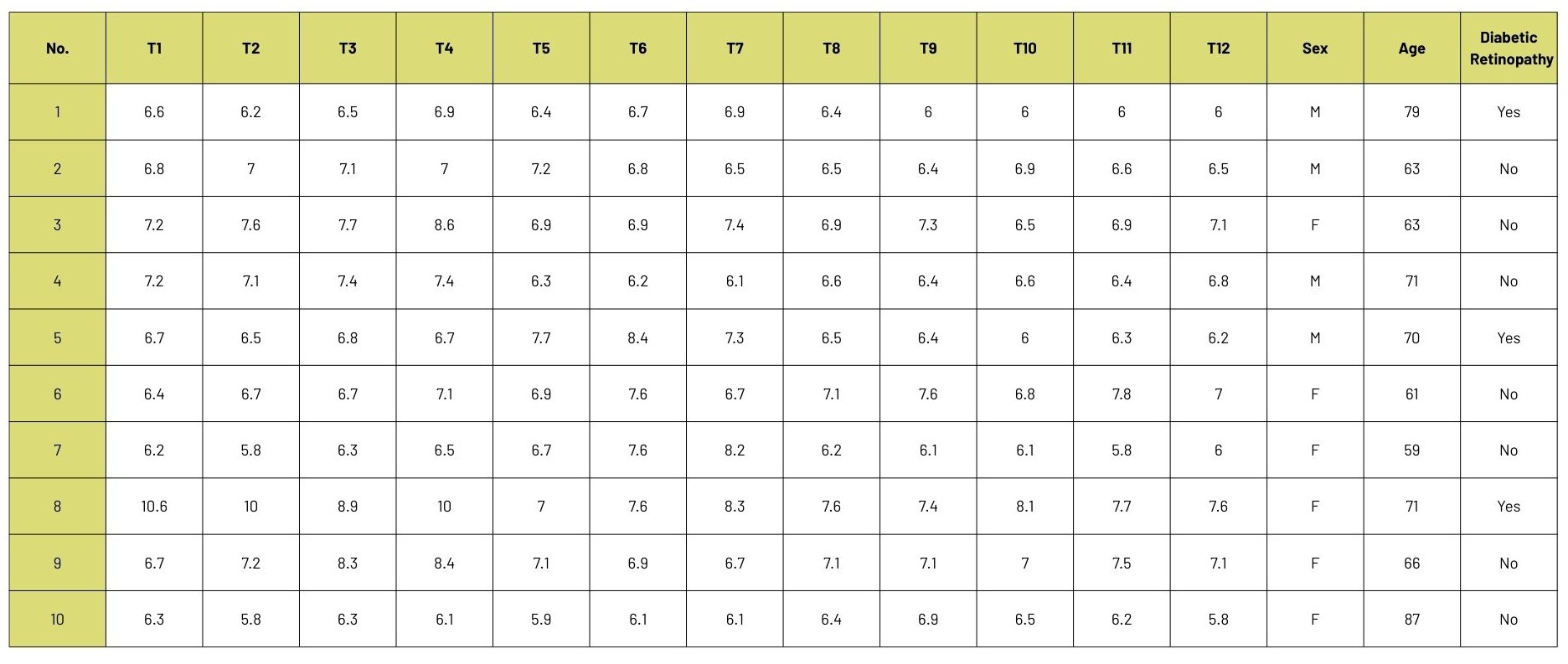}
    \caption{The first 10 rows of the dataset}
    \label{fig:10rows}
\end{figure}

The dataset of type 2 diabetes patients was retrospectively collected from Siriraj Hospital between 2015 and 2023. It comprises clinical records of diabetes patients who had visited the clinic at least 12 times, with the intervals between consecutive visits ranging between two and seven months. A total of 1,989 patients were eligible for the present study. The dataset includes laboratory test results, including HbA1c levels—crucial for assessing long-term glycemic control in diabetes patients—which were the focus of this study. Each patient has records of their HbA1c levels at 12 distinct time points, corresponding to their clinical visits. Other relevant clinical variables were also collected, including fasting blood glucose levels (FBS), cholesterol profiles (LDL, HDL, and triglycerides), kidney function parameters (creatinine and estimated glomerular filtration rate (eGFR)), diabetic retinopathy status, and demographic data (age and sex). However, due to the substantial proportion of missing values for these clinical variables, we only analyzed the diabetic retinopathy status and demographic background in addition to HbA1c. Figure \ref{fig:10rows} presents the records of the first 10 patients from the dataset.

\subsection{Diabetes patients segmentation} \label{diabetesapp}

Patients' medical laboratory results exemplify the type of dataset previously discussed. Using our clustering method, differences in the timings of two or more patients' clinical visits should not affect the outcome if the trends, patterns, and levels of their laboratory results are similar.  We applied 4TaStiC to differentiate diabetes patients, focusing on both their HbA1c levels and behaviors. Specifically, two patients with an almost identical blood glucose level but exhibiting different behaviors, as shown in Figure \ref{fig:ex1}, should be assigned to different groups. Conversely, two patients with slightly different glucose levels but exhibiting the same behavior should be assigned to the same group.

\begin{figure*}[hbt!]
\centering\includegraphics[width=12cm]{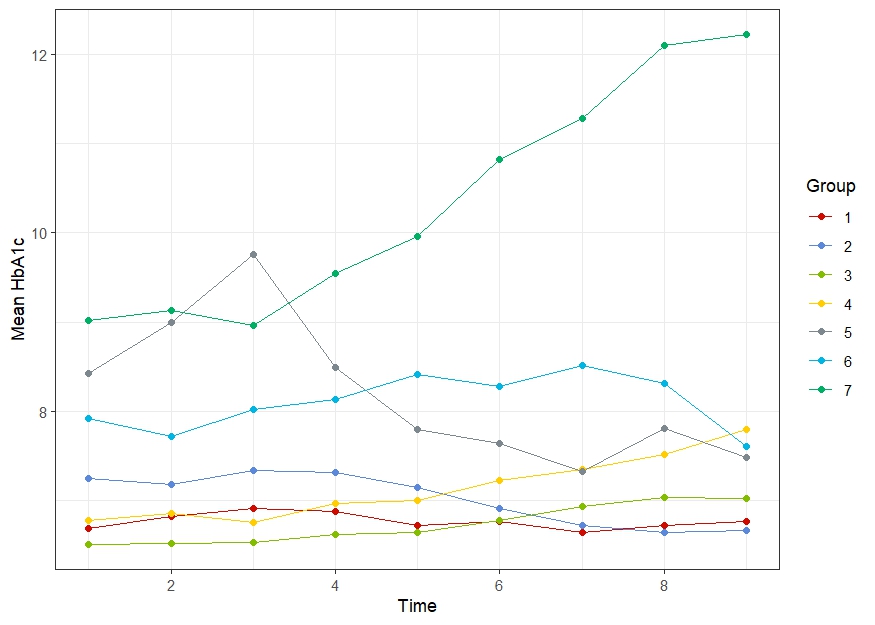}
\caption{Mean HbA1c of each visit from the best matched nine visits from each group.}
\label{fig:meanHbA1c}
\end{figure*}

\begin{table}[h]
\centering
\includegraphics[width=12cm]{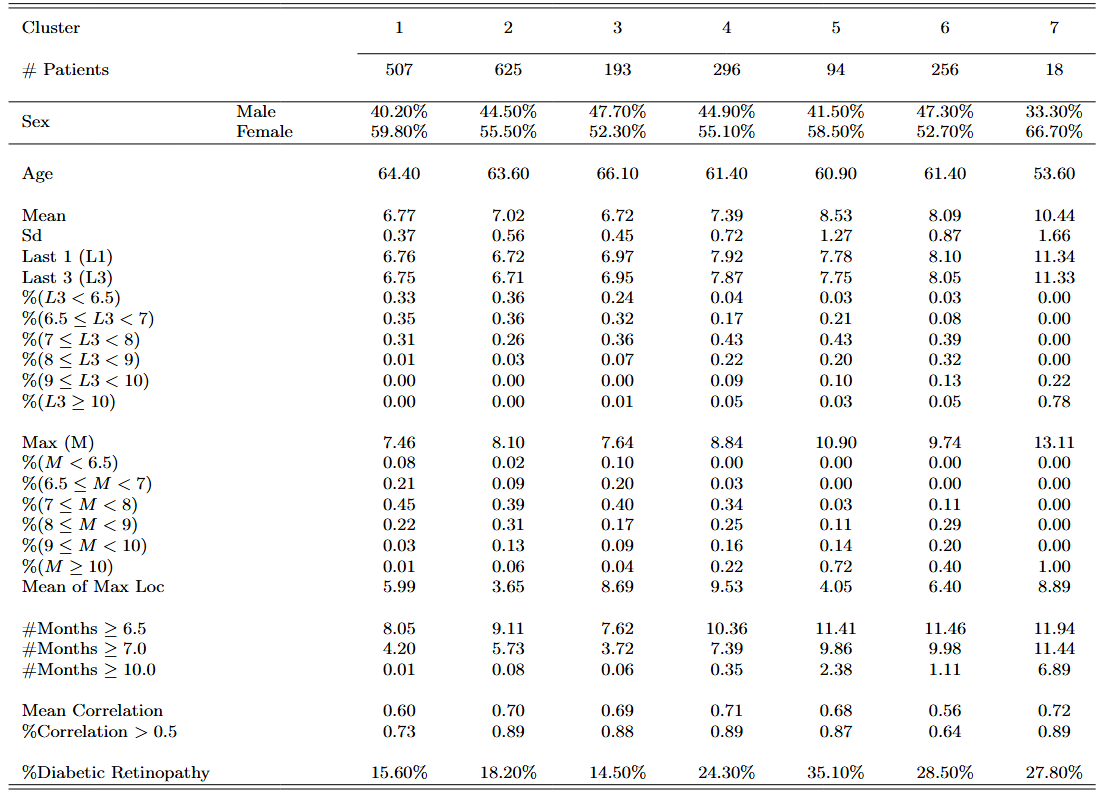}
\caption{Diabetes patients' characteristics in each group. HbA1c values are in percentages.}
\label{tab:characteristic}
\end{table}

\begin{figure*}[h]
\centering\includegraphics[width=12cm]{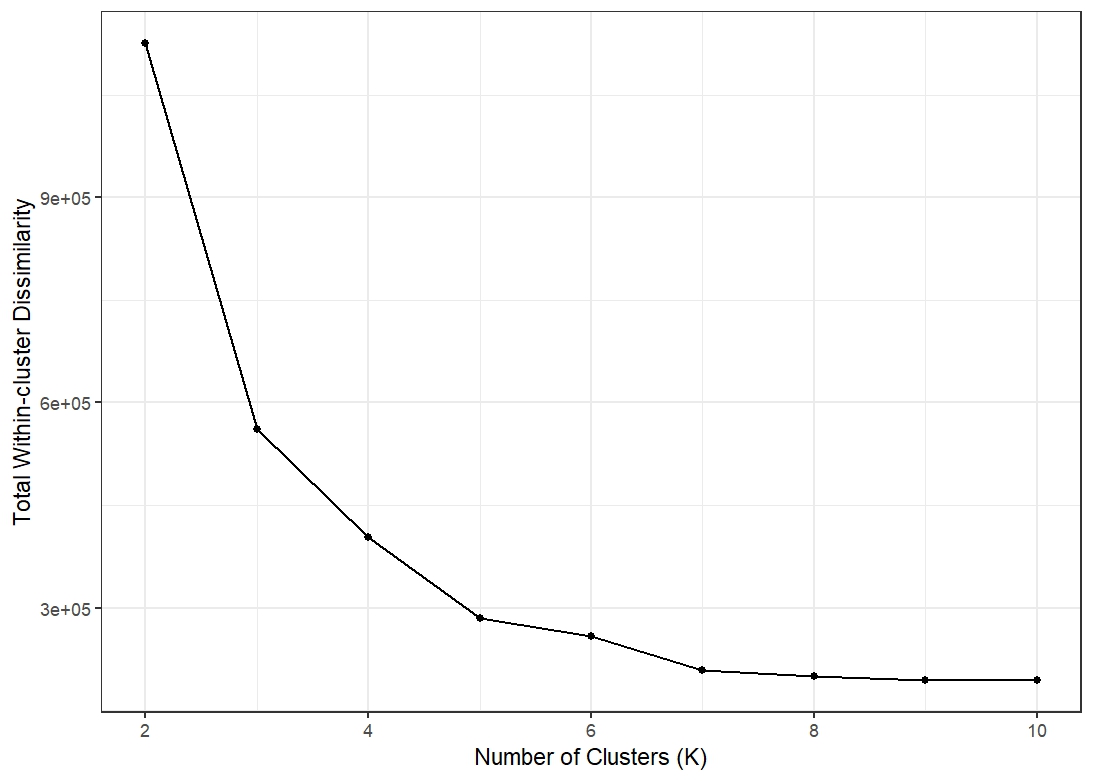}
\caption{Elbow method based on the 4TaStiC within cluster dissimilarity}
\label{fig:application_elbow}
\end{figure*}

To apply our 4TaStiC to the segmentation of diabetes patients, we set the parameters as follows: $\alpha$ is the same as in \eqref{alpha}, with $p = 0.09$, $L = 3$, $E = \{-0.075,0,0.075\}$ and $C= 0$. This selection is explained in Subsection \ref{performance} when testing performance using artificial datasets. Then, we perform the elbow method based on the 4TaStiC within-cluster dissimilarity to select the final number of clusters. The plot for $K$ from $2$ to $10$ is shown in Figure \ref{fig:application_elbow}.  We intentionally limit the number of clusters to less than 10 to facilitate a more concise and interpretable summary of our findings. It should be noted, however, that for practical implementation, a greater value of $K$ may be appropriate to capture patient characteristics. Considering visible elbows at $K = 3,5$, and $7$, we select $K=7$ as the final number of groups to explore the different detailed characteristics. Table \ref{tab:characteristic} summarizes the characteristics of patients in each group.

Table \ref{tab:characteristic} presents patient groups stratified according to the mean HbA1C level of the last three visits and the mean maximum, demonstrating three slightly high, three moderately high, and one extremely high HbA1c group. These seven groups each contain 507, 625, 193, 296, 94, 256, and 18 patients, respectively. Groups with similar HbA1c levels can be differentiated by their distinct patterns,  identified through their within-cluster correlations using the time and trend traveling technique. These differences are visualized in Figure \ref{fig:meanHbA1c}, which displays the mean HbA1c across the nine best time traveling matched visits from 30 patients randomly selected from each group (with all 18 patients plotted for the smallest group). We only plotted nine visits because we applied the time traveling technique with a maximum shift of $L=3$ from the total time step of $T=12$ visits, therefore nine is the minimum number of time steps used for computing the dissimilarity.

The slightly high HbA1c groups are classified as stable, slightly increasing, or slightly declining. The moderately high groups can be roughly categorized as increasing, declining, and quite stable. The extremely high HbA1c group exhibits a dramatic increasing trend. The similar patterns within each group are confirmed by the average within-cluster correlation with trend traveling, which ranges from 0.56 to 0.72. In all seven groups, as again shown in Table \ref{tab:characteristic}, the percentage of pairs with a correlation of 0.5 or above is 64\% or more and as high as 87\% and above in groups 2, 3, 5, 6, and 7.

We also present each group's in-depth characteristics in Table \ref{tab:characteristic}, namely, how many of the 12 visits record an HbA1c greater than or equal to 6.5\%, 7.0\%, and 10.0\%, and the percentages of patients whose max HbA1c result and mean of the most recent three HbA1c results fall within the following intervals: $(0,6.5)$, $[6.5,7.0)$, $[7.0,8.0)$, $[8.0,9.0)$, $[9.0,10.0)$, and $[10.0,\infty)$. We observe that all patients in the extremely high HbA1c group have an average HbA1c of above 9\% for their last three visits. The average HbA1c of most patients in the slightly and moderately high groups is below 8\% and ranges from 7\% to 9\%, respectively. These characteristics suggest, for instance, that Group 4 has recently been engaging in more risky behavior, even though their mean HbA1c is lower than that of Group 5. The mean of the maximum location can tell us the approximate worst period for patients in each group. It is clear that Groups 2 and 4 have better results than Groups 4 and 7 during the most recent visits. We can also see that there is a higher percentage of males in the critical group than in the other groups, but the group size is small.

Finally, we consider the percentage of patients in each group who have been diagnosed with diabetic retinopathy. The results are quite intuitive. Group 5, which at one point in the past had very high HbA1c levels, includes the highest percentage of patients (35.10\%) with diabetic retinopathy. It is possible that this group had a history of even higher HbA1c before the 12 visits captured in our dataset. Groups 4 and 7, whose HbA1c levels exhibit increasing trends, have lower proportions of complications at 24.30\% and 27.80\%, respectively. These two groups are at risk of additional complications in the future, especially Group 7, whose patients have extremely high HbA1c levels—without prevention measures, the percentage of patients with diabetic retinopathy may increase to around 35\%, as is the case in Group 5. In Group 6, 28.5\% of the patients have complications, which is reasonable given this group is similar to Group 5 but more stable. Between 14.5\% and 18.2\% of the patients in Groups 1 to 3, who have slightly high HbA1c, are diagnosed with complications.

\subsection{Diabetes patients' groups summary and medical implementation} \label{sec:groupsum}

Here, we discuss the characteristics of patients in each cluster from the previous subsection. According to Table \ref{tab:characteristic} and Figure \ref{fig:meanHbA1c}, each group's characteristics can be summarized as follows.

{\bf Group 1 (Looking good!):} This is a group of 507 patients with a slightly high average HbA1c over their last 12 visits, who are on a stable trend with small variation. This group represents individuals who have a history of mild diabetes and maintain their conditions well. These patients may be recommended to keep their diets consistent, exercise, and to keep up with what they have been doing.

{\bf Group 2 (Keep up your great work!):} This is a group of 625 patients whose HbA1c levels were moderately high at the first visit and then decreased until they were even lower than those of Group 1 by the 12th visit. This group represents individuals who have a history of diabetes but have now become more stable. They may be recommended to maintain their diets, exercise, and keep up with what they have been doing.

{\bf Group 3 (Still Okay but getting worse!):} This is a group of 193 patients with slightly high HbA1c levels at the beginning of the records. However, the increasing trend during the last few visits suggests that some risks have recently developed. This group may be recommended to slightly adjust their diets and attempt more exercise before it is too late.

{\bf Group 4 (Getting much riskier!):} This is a group of 296 patients whose HbA1c levels are currently on increasing trends. For some time, they had a slightly high HbA1c, before later developing risk. This group may be recommended to put more effort into their diets and exercise. The doctor may also prescribe new medication to prevent the worsening of their diabetes. This group must be carefully monitored for complications such as diabetic retinopathy.

{\bf Group 5 (Keep up your hard work, and try harder!):} This is a group of 94 patients with very high HbA1c levels at the first visit. However, the latest treatments seem to have been effective, and their HbA1c decreased rapidly after reaching very high peaks. This group may be recommended to maintain their prescriptions, and to strictly keep up with their diet and exercise regimens.

{\bf Group 6 (Try to be consistent!):} This is a group of 256 patients with moderately high HbA1c. Their HbA1c levels were very high and have decreased recently. This group may be recommended to keep up with their diets, exercise, and keep up with what they are currently doing. They should try to be more consistent in their behaviors to avoid worsening again. If new medication was prescribed recently, they seem to be effective.

{\bf Group 7 (Not looking good!):} This is a group of 18 patients with extremely high HbA1c levels. They used to have the same levels as Group5. However, their HbA1c levels have increased rapidly over the 12 visits, and they have now become the most critical patients. This group may be recommended to put serious effort into maintaining a healthy diet and getting exercise. The doctor may also prescribe new medication to prevent severe consequences. This group must be very closely monitored for complications such as diabetic retinopathy.

\begin{figure*}[hp]
\centering
\resizebox{1.0\textwidth}{!}{%
\begin{tabular}{c}
\centering\includegraphics[width=8cm]{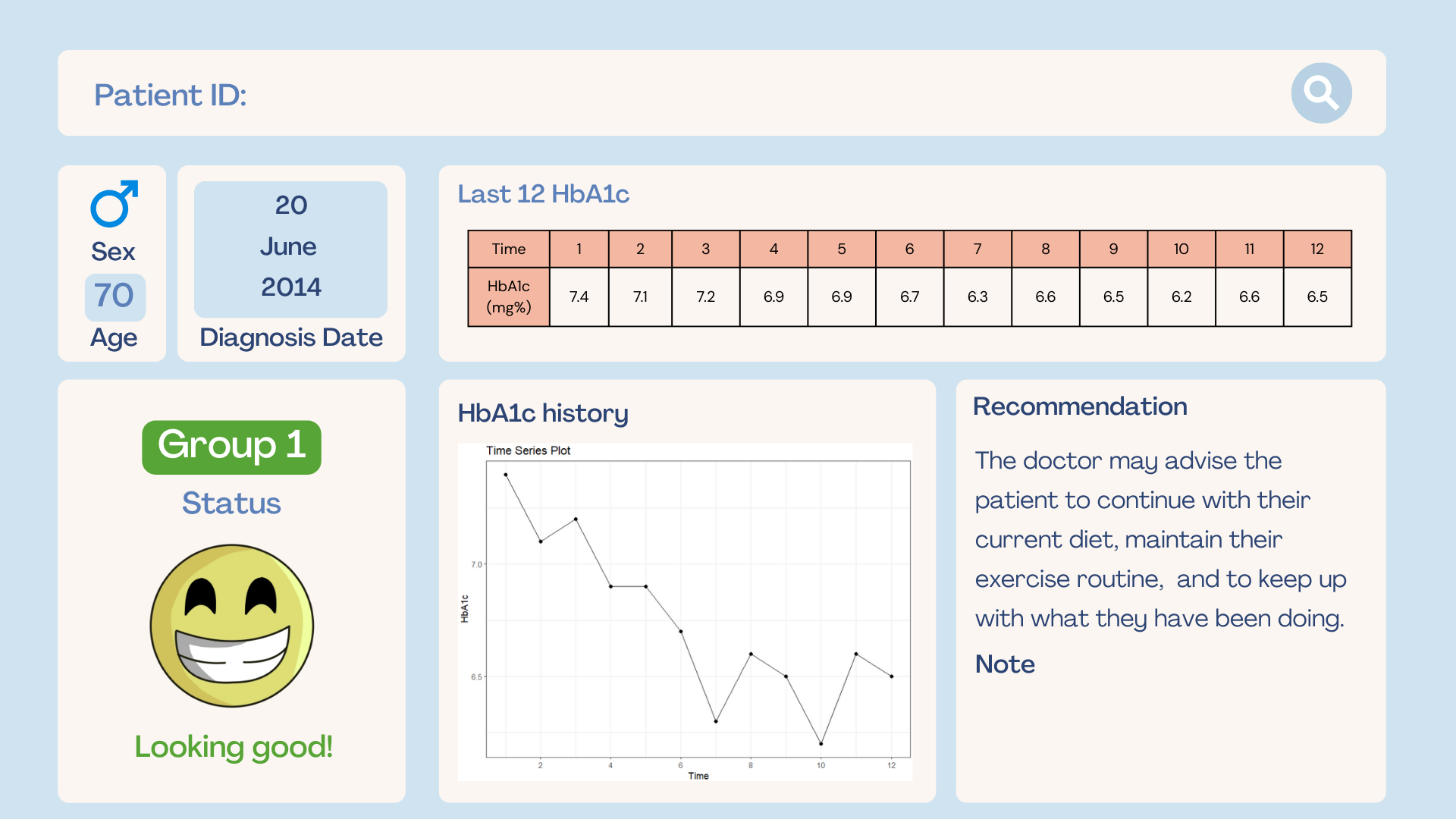}
\centering\includegraphics[width=8cm]{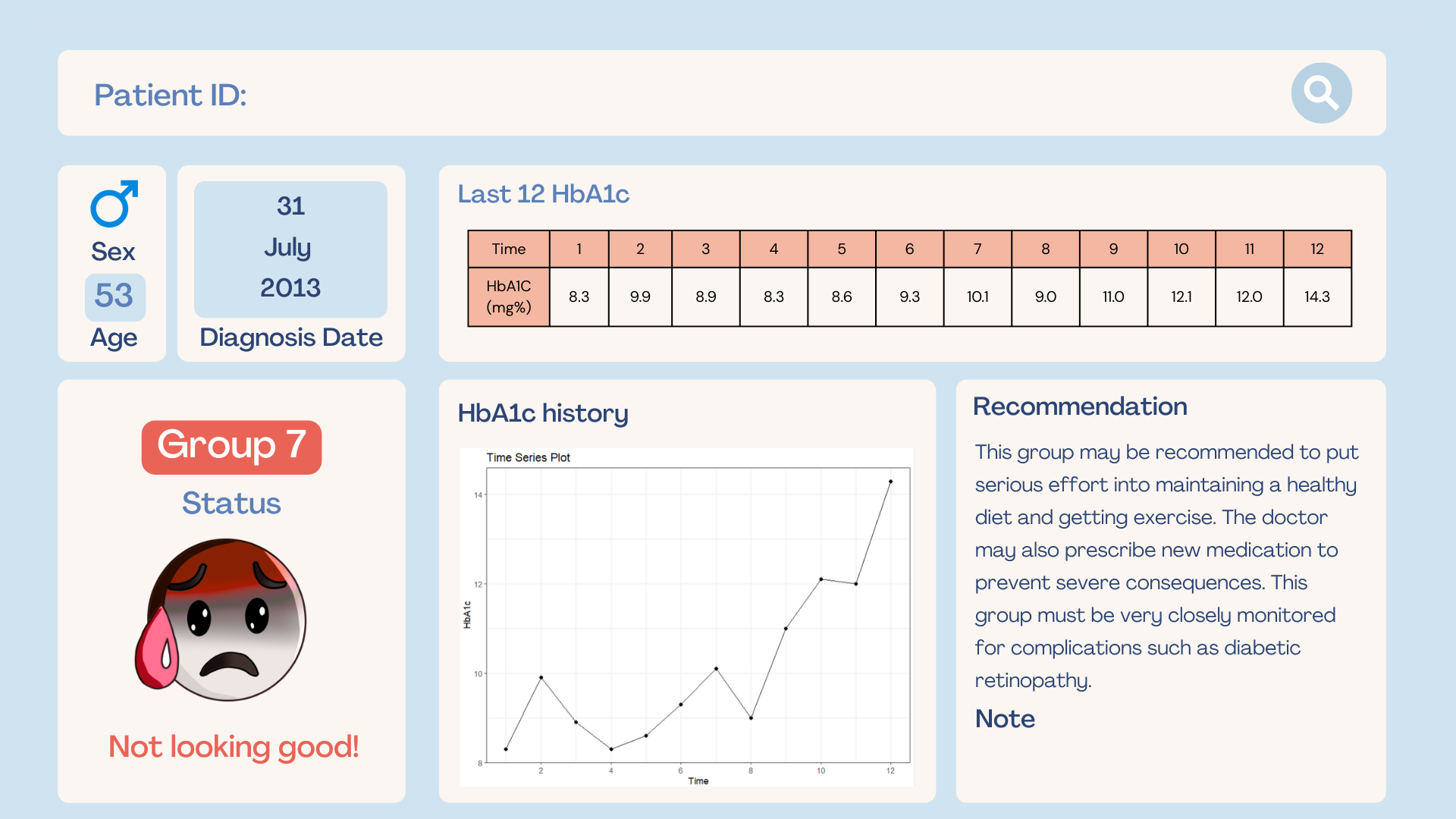} 
\end{tabular}
}
\caption{An example of dashboards for Group1 and Group7}
\label{dashboard}
\end{figure*}

To implement our method in clinical settings, it would be beneficial to develop a built-in dashboard synced with our learning model to present the doctor with each patient's status to enable the doctor to recognize patient behavior over time without the need to review the entire laboratory time series. In a real scenario, we may use a larger number of groups and longer time steps if available, as these can show more specific characteristics, such as a group of patients who only control themselves when their laboratory results reach a certain critical point. Figure \ref{dashboard} shows an example of dashboards for Group1 and Group7.

\section{Conclusion} \label{sec:discuss}

In this work, we introduce a time series clustering algorithm, 4TaStiC, which contributes to the area where time points are flexible and time series patterns are important. We also develop an R package ``FourTaStiC'' available on \url{https://github.com/nwiroonsri/FourTaStiC}. The summary of the method is presented in the following subsection. The performance of 4TaStiC is tested and compared with existing algorithms, including K-means, hierarchical clustering with Euclidean distance, Pearson correlation dissimilarity, cross-correlation dissimilarity, DTW, GAK, and LPWC on artificial datasets. 4TaStiC outperforms the existing methods in the aspect we intended. Our proposed method is then applied to classify 1,989 type 2 diabetes patients from Siriraj Hospital into seven groups, summarized in Subsection \ref{sec:groupsum}. This certainly benefits doctors in making efficient recommendations using information from the entire patient's time series data.

\subsection{4TaStiC summary and when to use it}

The proposed 4TaStiC is very flexible, allowing users to select the time and trend traveling parameters. We note that time traveling works only when data points with different timelines should be considered the same and handled similarly, such as patients' lab results, athletes' efficiency statistics by game, etc. It should not be applied to other fields where the time points must be fixed due to seasonality, such as monthly electricity consumption or agricultural production, etc. The trend traveling is integrated into 4TaStiC to maximize the opportunity to group data points with similar patterns and slightly different trends; however, this parameter is sensitive to small changes, and it must be ensured that data points rotating by that small angle can be meaningfully classified into the same group. The only main drawback of our proposed method is its computational time since it requires computing distances between all pairs of data points before and after the time and trend traveling is applied.

Figure \ref{fig:ex1} illustrates our idea, showing two diabetes patients who visit the doctor at different times with somewhat different durations between visits. Their lab results' patterns and levels are similar, and only slightly different. We may say that, for the bottom plots, the two patients' HbA1c tended to increase at the beginning of the record and have been decreasing after exceeding some high points. This could be because they have maintained healthy diets or were prescribed effective medicine once their HbA1c level rose too high.

\subsection{Future works}

Future research will include developing and analyzing better methods for detecting the number of clusters, which will allow us to detect the final number of clusters more accurately. Currently, the elbow method provides several elbows, but sometimes does not include the correct one. We will also apply 4TaStiC to density-based clustering, such as DBSCAN and OPTICS, which are compatible with more versatile types of data and allow outliers. It is also worth considering the application of 4TaStiC to different fields, such as sport science, finance, and social science. Implementing 4TaStiC with diabetes patients' data in the form of a web application or a dashboard is also of interest.

\section*{Acknowledgment}
Both authors appreciate Siriraj Hospital for allowing us to access the data of their diabetes patients. They are grateful for fruitful conversations with the team of Siriraj Population Health and Nutrition Research Group (SPHERE), Research Department, Faculty of Medicine Siriraj: Korapat Mayurasakorn, Phongthana Pasookhush, Pichanun Mongkolsucharitkul, Apinya Surawit, and Sureeporn Pumeiam regarding the medical implementation. O.P. would like to thank Petchra Pra Jom Klao Master’s Degree Research Scholarship from King Mongkut’s University of Technology Thonburi under Grant number: 32/2567. She is especially thankful to Sutthipong Sindee for the valuable discussion to this work. N.W. would like to thank KMUTT Partnering Initiative Grant, fiscal year 2024 under KIRIM number 28105 for research financial support.

\bigskip
{\bf Contributions}
N.W. contributed to introducing 4TaStiC, exploring its properties, and the overall experimental design. O.P. contributed to coding, preparing the R package, performing the experiments, and creating graphics. Both authors contributed to writing and editing the manuscript.

\bigskip\noindent
{\bf Funding} This research project was supported by King Mongkut's University of Technology Thonburi: KMUTT Partnering Initiative Grant, fiscal year 2024 under KIRIM number 28105.

\bigskip\noindent
{\bf Data Availability} The artificial datasets used in this study can be obtained from \url{https://github.com/nwiroonsri/FourTaStiC}. The diabetes dataset is confidential.

\bigskip\noindent
{\bf Ethics approval} The study protocol was approved by the Ethics Committee of the Human Research Protection Unit, Faculty of Medicine Siriraj Hospital, Mahidol University board (COA no. Si 661/2024), and KMUTT-IRB-COE-2025-009.

\bigskip\noindent
{\bf Conflict of interest} The authors declare no competing interests.




\begin{thebibliography}{10}
\providecommand{\url}[1]{{#1}}
\providecommand{\urlprefix}{URL }
\providecommand{\doi}[1]{\url{https://doi.org/#1}}

\bibitem{khan2020}
M.A.B. Khan, M.J. Hashim, J.K. King, R.D. Govender, H.~Mustafa, J.~Al~Kaabi, Epidemiology of type 2 diabetes - global burden of disease and forecasted trends.
\newblock Journal of Epidemiology and Global Health \textbf{10}(1), 107--111 (2020).
\newblock \doi{10.2991/jegh.k.191028.001}

\bibitem{WHO2024}
{World Health Organization}.
\newblock Urgent action needed as global diabetes cases increase four-fold over past decades (2024).
\newblock \urlprefix\url{https://www.who.int/news/item/13-11-2024-urgent-action-needed-as-global-diabetes-cases-increase-four-fold-over-past-decades}

\bibitem{DR2}
N.~Puangmee, Nursing of diabetic retinopathy in type 2 diabetes patients.
\newblock Kuakarun Journal of Nursing \textbf{25}(1), 217–227 (2018).
\newblock \urlprefix\url{https://he01.tci-thaijo.org/index.php/kcn/article/view/131951}

\bibitem{DR2_prevent}
S.G. Schorr, H.P. Hammes, U.A. M{\"u}ller, H.H. Abholz, R.~Landgraf, B.~Bertram, The prevention and treatment of retinal complications in diabetes.
\newblock Dtsch. Arztebl. Int. \textbf{113}(48), 816--823 (2016)

\bibitem{diabetes_het}
R.D. Leslie, R.C.W. Ma, P.W. Franks, K.J. Nadeau, E.R. Pearson, M.J. Redondo, Understanding diabetes heterogeneity: key steps towards precision medicine in diabetes.
\newblock The Lancet Diabetes \& Endocrinology \textbf{11}(11), 848--860 (2023).
\newblock \doi{10.1016/S2213-8587(23)00159-6}.
\newblock 

\bibitem{precisionmed1}
R.B. Prasad, L.~Groop, Precision medicine in type 2 diabetes.
\newblock J. Intern. Med. \textbf{285}(1), 40--48 (2019).
\newblock \doi{10.1111/joim.12859}

\bibitem{ThaiHS}
{World Health Organization. Regional Office for the Western Pacific}, {Asia Pacific Observatory on Health Systems and Policies}, \emph{Thailand health system review}, vol.~13 (WHO Regional Office for the Western Pacific, Manila, Philippines, 2024)

\bibitem{ThaiHS2}
N.~Kakandee, A.~Cheevakasemsook, D.~Triwichitkhun, Job burnout of generation y professional nurses at a government hospital.
\newblock Journal of The Royal Thai Army Nurses \textbf{21}(1), 293–301 (2020).
\newblock \urlprefix\url{https://he01.tci-thaijo.org/index.php/JRTAN/article/view/241573}

\bibitem{SI}
Siriraj hospital -- thailand's excellent medical hub - {SIRIRAJ}.
\newblock \url{https://www2.si.mahidol.ac.th/en/news-events/siriraj-hospital-thailands-excellent-medical-hub/} (2022)

\bibitem{clustering}
C.~Sammut, G.I. Webb (eds.), \emph{Clustering} (Springer US, Boston, MA, 2010), pp. 180--180.
\newblock \doi{10.1007/978-0-387-30164-8_124}.


\bibitem{bio1}
J.~Li, S.~Chen, X.~Pan, Y.~Yuan, H.B. Shen, Cell clustering for spatial transcriptomics data with graph neural networks.
\newblock Nature Computational Science \textbf{2}(6), 399--408 (2022)

\bibitem{bio2}
A.~Dominguez~Mantes, D.~Mas~Montserrat, C.D. Bustamante, X.~Gir{\'o}-i Nieto, A.G. Ioannidis, Neural admixture for rapid genomic clustering.
\newblock Nature computational science \textbf{3}(7), 621--629 (2023)

\bibitem{xue2017joint}
L.~Xue, Y.~Liu, Z.Q. Gu, Z.H. Li, X.P. Guan, Joint design of clustering and in-cluster data route for heterogeneous wireless sensor networks.
\newblock International Journal of Automation and Computing \textbf{14}(6), 637--649 (2017)

\bibitem{xu2017novel}
D.G. Xu, P.L. Zhao, C.H. Yang, W.H. Gui, J.J. He, A novel minkowski-distance-based consensus clustering algorithm.
\newblock International Journal of Automation and Computing \textbf{14}(1), 33--44 (2017)


\bibitem{timeclust}
T.~{Warren Liao}, Clustering of time series data—a survey.
\newblock Pattern Recognition \textbf{38}(11), 1857--1874 (2005).
\newblock \doi{https://doi.org/10.1016/j.patcog.2005.01.025}.


\bibitem{elecmonitoring}
X.~Yu, L.~Lu, J.~Qi, Y.~Qian, L.~Zhao, C.~Tan, Y.~Chen, Z.~Han, A clustering fractional-order grey model in short-term electrical load forecasting.
\newblock Scientific Reports \textbf{15}(1), 6207 (2025)

\bibitem{powerconsumption}
K.~Alsalem, A hybrid time series forecasting approach integrating fuzzy clustering and machine learning for enhanced power consumption prediction.
\newblock Scientific Reports \textbf{15}(1), 6447 (2025)

\bibitem{timefinance}
F.~Pattarin, S.~Paterlini, T.~Minerva, Clustering financial time series: an application to mutual funds style analysis.
\newblock Computational Statistics \& Data Analysis \textbf{47}(2), 353--372 (2004)

\bibitem{timepattern}
J.~Lin, E.~Keogh, L.~Wei, S.~Lonardi, Experiencing {SAX}: a novel symbolic representation of time series.
\newblock Data Min. Knowl. Discov. \textbf{15}(2), 107--144 (2007)

\bibitem{ex_timeclust1}
J.~Qiu, Y.~Hu, L.~Li, A.M. Erzurumluoglu, I.~Braenne, C.~Whitehurst, J.~Schmitz, J.~Arora, B.A. Bartholdy, S.~Gandhi, P.~Khoueiry, S.~Mueller, B.~Noyvert, Z.~Ding, J.N. Jensen, J.~de~Jong, Deep representation learning for clustering longitudinal survival data from electronic health records.
\newblock Nat. Commun. \textbf{16}(1), 2534 (2025).
\newblock \doi{10.1038/s41467-025-56625-z}

\bibitem{ex_timeclust2}
H.~Saito, H.~Yoshimura, K.~Tanaka, H.~Kimura, K.~Watanabe, M.~Tsubokura, H.~Ejiri, T.~Zhao, A.~Ozaki, S.~Kazama, M.~Shimabukuro, K.~Asahi, T.~Watanabe, J.J. Kazama, Predicting {CKD} progression using time-series clustering and light gradient boosting machines.
\newblock Sci. Rep. \textbf{14}(1), 1723 (2024).
\newblock \doi{10.1038/s41598-024-52251-9}

\bibitem{ex_timeclust3}
M.T. Bahadori, Z.C. Lipton, Temporal-clustering invariance in irregular healthcare time series.
\newblock arXiv preprint arXiv:1904.12206  (2019)

\bibitem{ex_timeclust4}
S.V. Bhavani, L.~Xiong, A.~Pius, M.~Semler, E.T. Qian, P.A. Verhoef, C.~Robichaux, C.M. Coopersmith, M.M. Churpek, Comparison of time series clustering methods for identifying novel subphenotypes of patients with infection.
\newblock Journal of the American Medical Informatics Association \textbf{30}(6), 1158--1166 (2023).
\newblock \doi{10.1093/jamia/ocad063}.


\bibitem{mpox}
V.~Borges, M.P. Duque, J.V. Martins, P.~Vasconcelos, R.~Ferreira, D.~Sobral, A.~Pelerito, I.L. de~Carvalho, M.S. N{\'u}ncio, M.J. Borrego, et~al., Viral genetic clustering and transmission dynamics of the 2022 mpox outbreak in portugal.
\newblock Nature Medicine \textbf{29}(10), 2509--2517 (2023)

\bibitem{dtw1}
V.~Velichko, N.~Zagoruyko, Automatic recognition of 200 words.
\newblock International Journal of Man-Machine Studies \textbf{2}(3), 223--234 (1970).
\newblock \doi{https://doi.org/10.1016/S0020-7373(70)80008-6}.


\bibitem{dtw2}
H.~Sakoe, S.~Chiba, Dynamic programming algorithm optimization for spoken word recognition.
\newblock IEEE Transactions on Acoustics, Speech, and Signal Processing \textbf{26}(1), 43--49 (1978).
\newblock \doi{10.1109/TASSP.1978.1163055}

\bibitem{gak}
M.~Cuturi, \emph{Fast global alignment kernels} (Omnipress, Madison, WI, USA, 2011), ICML'11, p. 929–936

\bibitem{autocorr}
A.~Egri, I.~Horváth, F.~Kovács, R.~Molontay, K.~Varga, \emph{Cross-correlation based clustering and dimension reduction of multivariate time series}, in \emph{2017 IEEE 21st International Conference on Intelligent Engineering Systems (INES)} (2017), pp. 000241--000246.
\newblock \doi{10.1109/INES.2017.8118563}

\bibitem{lpwc}
T.~Chandereng, A.~Gitter, Lag penalized weighted correlation for time series clustering.
\newblock BMC Bioinformatics \textbf{21}(1), 21 (2020).
\newblock \doi{https://doi.org/10.1186/s12859-019-3324-1}

\bibitem{hc1}
R.~Sibson, Slink: An optimally efficient algorithm for the single-link cluster method.
\newblock The Computer Journal \textbf{16}(1), 30--34 (1973).
\newblock \doi{10.1093/comjnl/16.1.30}.


\bibitem{hc2}
D.~Defays, An efficient algorithm for a complete link method.
\newblock The Computer Journal \textbf{20}(4), 364--366 (1977).
\newblock \doi{10.1093/comjnl/20.4.364}.
\newblock 

\bibitem{DBSCAN}
M.~Ester, H.P. Kriegel, J.~Sander, X.~Xu, et~al., \emph{A density-based algorithm for discovering clusters in large spatial databases with noise}, in \emph{kdd}, vol.~96 (1996), pp. 226--231

\bibitem{OPTICS}
M.~Ankerst, M.M. Breunig, H.P. Kriegel, J.~Sander, Optics: Ordering points to identify the clustering structure.
\newblock ACM Sigmod record \textbf{28}(2), 49--60 (1999)

\bibitem{RDPC2025}
C.~Charoensuk, N.~Wiroonsri.
\newblock Ranked differences pearson correlation dissimilarity with an application to electricity users time series clustering (2025).
\newblock \urlprefix\url{https://arxiv.org/abs/2505.02173}

\bibitem{RStudio}
{RStudio Team}, \emph{RStudio: Integrated Development Environment for R}.
\newblock RStudio, PBC., Boston, MA (2020)

\bibitem{dtwclust}
A.~Sardá-Espinosa, Time-series clustering in r using the dtwclust package.
\newblock The R Journal  (2019).
\newblock \doi{10.32614/RJ-2019-023}

\bibitem{Rlpwc}
T.~Chandereng, A.~Gitter, \emph{LPWC: Lag Penalized Weighted Correlation for Time Series Clustering} (2020).
\newblock R package version 1.0.0

\bibitem{NCI}
N.~Wiroonsri, Clustering performance analysis using a new correlation-based cluster validity index.
\newblock Pattern Recognition \textbf{145}, 109910 (2024).
\newblock \doi{https://doi.org/10.1016/j.patcog.2023.109910}.


\bibitem{BayesCVI}
O.~Preedasawakul, N.~Wiroonsri, A bayesian cluster validity index.
\newblock Computational Statistics \& Data Analysis \textbf{202}, 108053 (2025).
\newblock \doi{https://doi.org/10.1016/j.csda.2024.108053}.
\newblock 

\bibitem{Kmeans1}
S.P. Lloyd, Least squares quantization in pcm.
\newblock IEEE Trans. Inf. Theory \textbf{28}, 129--136 (1982).
\newblock \urlprefix\url{https://api.semanticscholar.org/CorpusID:10833328}

\bibitem{kmeans2}
J.~MacQueen, \emph{Some methods for classification and analysis of multivariate observations} (1967).
\newblock \urlprefix\url{https://api.semanticscholar.org/CorpusID:6278891}

\bibitem{ARI}
W.M. Rand, Objective criteria for the evaluation of clustering methods.
\newblock Journal of the American Statistical Association \textbf{66}(336), 846--850 (1971).
\newblock \doi{10.1080/01621459.1971.10482356}.

\end{thebibliography}


\end{document}